\documentclass[12pt, draftclsnofoot, onecolumn]{IEEEtran}
\usepackage[T1]{fontenc}

\usepackage[nocomma]{optidef}
\usepackage[ruled,vlined]{algorithm2e}
\usepackage{enumerate}
\usepackage{booktabs}

\usepackage[square, comma, compress, numbers]{natbib}

\ifCLASSINFOpdf
\else
\fi
\usepackage{amsmath}
\usepackage[cmintegrals]{newtxmath}
\usepackage{array}

\usepackage[font=footnotesize]{caption}
\usepackage{amsthm}
\newtheorem{theorem}{\textbf{Theorem}}

\newtheorem{lemma}{\textbf{Lemma}}
\newtheorem{remark}{\textbf{Remark}}

\usepackage{mathtools}
\newcommand\aseq{\stackrel{\mathclap{\normalfont\mbox{a.s.}}}{=}}

\usepackage[normalem]{ulem}
\usepackage{xcolor}

\newcommand{\added}[1]{{\leavevmode\color{black}#1}}
\newcommand{\wdh}[1]{{\leavevmode\color{black}#1}}
\usepackage{hhline}

\usepackage{enumitem}

\begin{document}
%
\title{Optimization of Mobile Robotic Relay Operation for Minimal Average Wait Time}
%
%
%

\author{Winston~Hurst,~\IEEEmembership{Student Member,~IEEE,}
        and~Yasamin~Mostofi,~\IEEEmembership{Fellow,~IEEE}
\thanks{Winston Hurst and Yasamin Mostofi are with the Department of Electrical and Computer Engineering, University of California, Santa Barbara, USA (email: \{winstonhurst, ymostofi\}@ece.ucsb.edu). This work was supported in part by NSF RI award 2008449.}
}

\maketitle
\vspace{-0.5in}
\begin{abstract}
This paper considers trajectory planning for a mobile robot which persistently relays data between pairs of far-away communication nodes. Data accumulates stochastically at each source, and the robot must move to appropriate positions to enable data offload to the corresponding destination. \added{The robot needs to minimize the average time that data waits at a source before being serviced. We are interested in finding optimal robotic routing policies consisting of 1) locations where the robot stops to relay (relay positions) and 2) conditional transition probabilities that determine the sequence in which the pairs are serviced. We first pose this problem as a non-convex problem that optimizes over both relay positions and transition probabilities.} To find approximate solutions, we propose a novel algorithm which alternately optimizes relay positions and transition probabilities. For the former, we find efficient convex partitions of the non-convex relay regions, then formulate a mixed-integer second-order cone problem. For the latter, we find optimal transition probabilities via sequential least squares programming. \added{We extensively analyze the proposed approach and mathematically characterize important system properties related to the robot’s long-term energy consumption and service rate. Finally, through extensive simulation with real channel parameters, we verify the efficacy of our approach.}

\end{abstract}

\vspace{-0.1in}
\begin{IEEEkeywords}
\vspace{-0.1in}
Autonomous robots, Unmanned Aerial Vehicle (UAV), relay systems, communication-aware robotics, UAV-assisted communication, polling systems
\end{IEEEkeywords}

%
\IEEEpeerreviewmaketitle

\vspace{-0.17in}
\section{Introduction}\label{sec:introduction}
\vspace{-0.05in}
%
%
%
%

\IEEEPARstart{S}{ignificant} advances in robotics over the past several years have created new possibilities in the design of communication systems. For example, unmanned vehicles \wdh{(ground or unmanned aerial vehicles)} may enable, extend, or improve networks via data muling \added{\cite{VFDTN, Kabir2011OptimalVO, Tsilomitrou2018MobileRT}}, relaying \added{\cite{carrillo2021, zhang2022, Park2019FlockingInspiredTP, reveliotis2022}}, or beamforming \added{\cite{MdharanMostofi_TCSN17,Evmorfos2022ReinforcementLF}}. Design of these mobility-enabled communication systems must account for both the motion and communication aspects of operation. \added{This field is referred to as \textit{communication-aware robotics}~\cite{ghaffarkhah2010channel,TRO19_MuralidharanMostofi, Autonomous19_MuralidharanMostofi, AR2020}. Muralidharan and Mostofi}~\cite{AR2020} provide a recent review of this area.

\begin{figure}[t]
\centering
\includegraphics[trim=0 20 0 0, clip, width=3.7in]{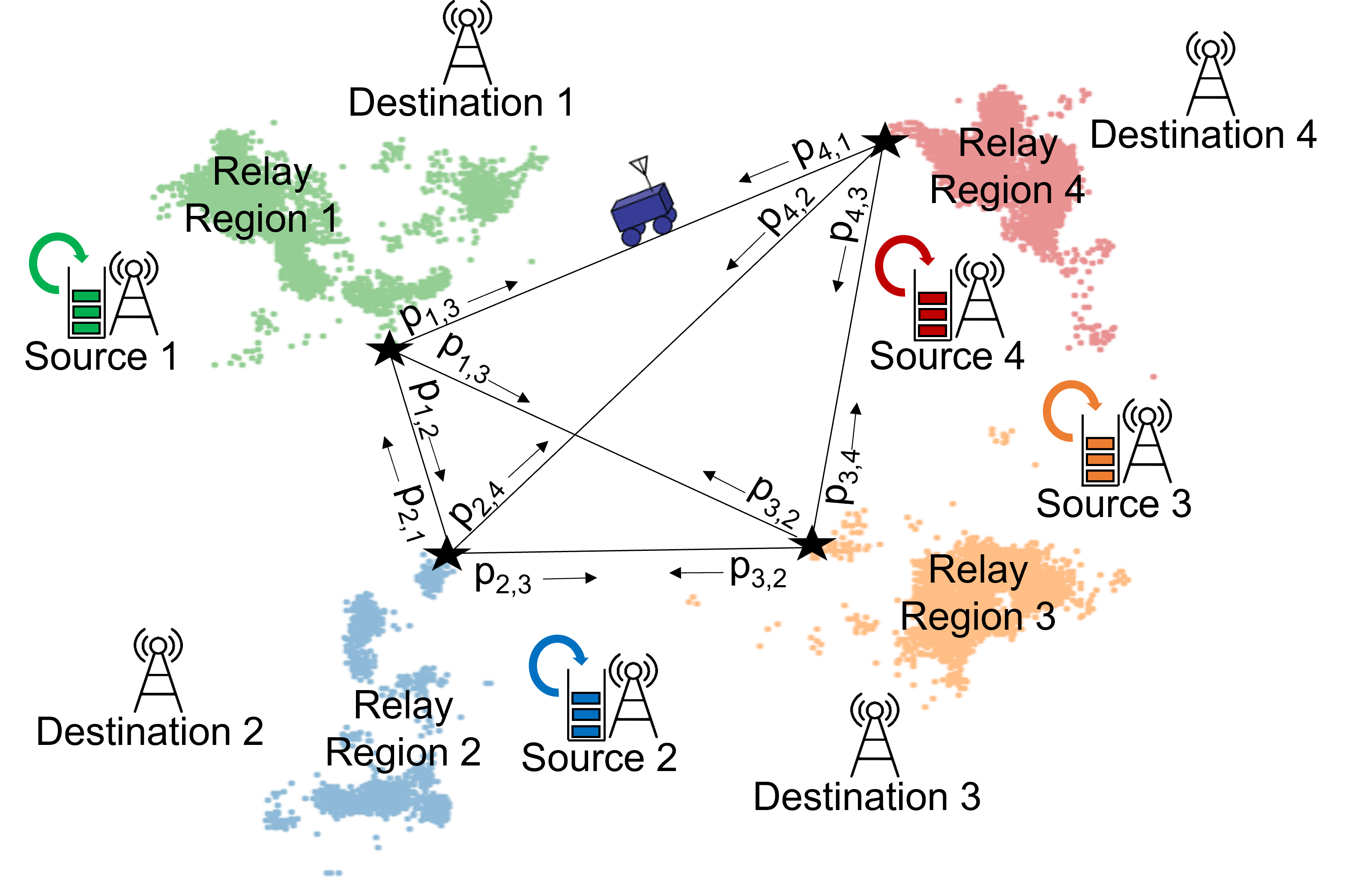}
\vspace{-0.1in}
\caption{At each source, data accumulates in a queue which must be offloaded to the corresponding destination while minimizing the average wait time. The relay \wdh{operation} is performed by a mobile robot, which needs to move to an optimum position within each feasible relay region to service the corresponding queue. After servicing queue $i$, the robot moves to service queue $j$ with probability $p_{i,j}$, which is found based on optimizing a stochastic robotic routing policy. See the color pdf for better viewing.}
\label{fig:overview}
\vspace{-0.27in}
\end{figure}

\added{In this paper, we consider the operation of a mobile robot which is tasked with persistently servicing several disparate communication links each consisting of a source and destination pair, as shown in Fig.~\ref{fig:overview}. Data arrives stochastically at each source and must be sent to its corresponding destination, which is too far away for direct communication. The mobile robot enables data transfer between each source-destination pair by creating a two-hop link between them. Examples of such systems include sensor networks \cite{henkel2008} and ad hoc networks deployed for search and rescue or after a natural disaster.}

\added{To effectively operate, the robot must identify regions, labeled \textit{Relay Regions} in Fig.~\ref{fig:overview}, for each source-destination pair where the link qualities from (to) the source (destination) are good enough to permit the robot to relay data from the source to the destination. Then, the robot must plan a trajectory which repeatedly visits these regions, allowing for persistent data transfer.}

\added{Several factors make this problem interesting and complex. First, in general, full channel information is not known to the robot. Therefore, the robot must accurately predict the relay regions with limited information. Doing so requires realistic modeling of highly non-convex spatial variations of communication channels, based only on sparse prior samples, as simplified models, \textit{e.g.}, disc models, may result in performance degradation. Second, as the real-world communication channels which determine the relay regions are irregular, the regions become non-convex and disjoint, making path planning through them challenging. Third, the robot may need to service some pairs more frequently than others due to potential heterogeneous data accumulation rates and asymmetry in spatial locations. Fourth, the path planning needs to be co-optimized with transitional routing probabilities. Finally, persistent operation calls for infinite horizon planning, requiring careful trajectory parameterization and resource constraint definitions.}

\added{To account for these factors, we employ a realistic, probabilistic channel estimation framework and consider stochastic trajectories parameterized by a set of relay positions (one per relay region) and transition probabilities, as shown in Fig.~\ref{fig:overview}. These \textit{stochastic robotic path planning policies} allow for arbitrary, non-homogeneous service frequencies~\cite{Srivastava2009} and provide security benefits in adversarial settings~\cite{grace2005}. Alternatively, they may be used as a basis for the construction of deterministic policies \cite{Boxma1993EfficientVO}, which may be difficult to directly optimize.}

\added{Utilizing mobile robots to relay data has received considerable attention in recent years~\cite{VFDTN, Kabir2011OptimalVO, evmorfos2022_frontiers, Mostofi_Murray_Secon04, GhaffarkhahMostofiACM13, Shen2016MobileRS, pogue2020, george2020}. In these problems, trajectory optimization considers paths which either move the robot from node to node (data muling) \cite{VFDTN, Kabir2011OptimalVO} or move the robot to locations where the wireless channel quality permits reliable communication \cite{Evmorfos2022ReinforcementLF, GhaffarkhahMostofiACM13, Shen2016MobileRS, pogue2020, george2020}. Closely related are persistent monitoring problems, in which a robot senses various locations in a workspace and transfers the sensed data to a remote station \cite{Ny2008ContinuousPP, khazaeni2015,Peters2019UAVSU, zhang2020, cassandras2022}, and communication-aware variations of the vehicle routing problem (VRP) \cite{celik2010, Luyo2020MixedIF}.}

\added{However, servicing a number of source-destination pairs with heterogeneous traffic, and in real-world channel environments, through visit location optimization and persistent path planning  differentiates this work from the literature. In much of existing work that addresses the robotic path planning component, either simplified channel models, \textit{e.g.}, disc models, are used~\cite{Park2019FlockingInspiredTP, Shen2016MobileRS, Ny2008ContinuousPP, celik2010}, or the robot visits a number of sites directly, \textit{i.e.}, without any communication component \cite{VFDTN, Kabir2011OptimalVO}.\footnote{Note that the term "relaying" in these papers refers to data muling (physically picking data up from one location and dropping it off at another) and differs from what relaying means in this paper.} This greatly simplifies optimization of visiting locations. Other work focuses on planning tours which visit each site exactly once~\cite{GhaffarkhahMostofiACM13, Peters2019UAVSU} rather than considering non-homogeneous visit frequencies. On the other hand, in most work that consider real communication issues, either path optimization from point A to point B is considered and/or only a single source-destination is assumed with a given starting point~\cite{AbdElmagid2019AveragePA, Shen2016MobileRS, pogue2020, george2020, Evmorfos2022ReinforcementLF,pogue2020}. Optimizing both the stop locations and the stochastic routing policy in realistic channel environments that result in highly non-convex relay regions, and while addressing multiple heterogeneous, persistent sources introduces exciting new challenges, which motivates the proposed mathematical framework of this paper.}

Our system of interest is also related to polling systems\wdh{, in which} a single server services multiple queues. In fact, we use seminal results which characterize average wait times for Markovian polling systems \cite{Boxma1989WaitingTI}. In fact, we show that polling system optimization becomes a special case of our problem of interest, and when appropriate, we draw analogies with polling system problems.

\added{However, such work is not in the context of robotics, so issues related to location optimization and realistic communication environments are not relevant. Even without our robotic-related issues, a survey of polling systems literature shows that finding general optimal operating policies is an open problem \cite{Perel2020APS, Vishnevsky2021UsingAM, Liu1992OnOP, Sato1997TheEO, Matveev2017NearoptimalityOS, Borst2018PollingPP}.  The additional considerations of location optimization and realistic communication environments adds further complexity to our problem.}

\wdh{We next summarize our contributions.}

\vspace{-0.04in}
\begin{enumerate}[leftmargin=1.5\parindent]
   \item \added{For the highly non-convex and intractable persistent relaying problem described above,} we introduce approximately-optimal robotic relay policies (AORP's) and bring a foundational understanding to this problem. Specifically, we propose a novel approach that iteratively minimizes the average wait time over both relay positions and the stochastic Markovian routing policy (i.e., robot transition probabilities from one service location to the next). \added{These stochastic policies} may be used as a basis for the construction of deterministic policies, as we shall also show.

    \item When optimizing the service locations, we show how each relay region can be predicted and efficiently partitioned into a set of convex regions. We then show how the optimum robot service locations can be found via an efficient mixed-integer second-order cone program (MISOCP) that minimizes the average wait time. We further show how the optimum Markovian robotic routing policy can be found via sequential least squares programming (SLSQP).

    \item Using a polling system model of the robot’s operation, we mathematically characterize the robot’s service time percentage, average power, and average service rate. As we shall see, our findings reveal interesting characteristics of the system in the long-term as well as the per stage context.

    \item We extensively test the proposed approach in realistic channel environments (channel parameters from real data) and show the impact of several different parameters on the performance of the system. We further compare with the state-of-the-art \added{and also validate our theoretical results}.
    \vspace{-0.05in}
\end{enumerate}

The rest of the paper proceeds as follows: Section~\ref{sec:modeling} introduces models for communication, channel prediction, \added{data accumulation and offloading,} and robot motion, along with a \added{Markov process model of the system}. Section~\ref{sec:problem} formalizes the minimal average wait time problem, and in Section~\ref{sec:AORP}, we propose a novel algorithm to find approximately-optimal solutions. Section~\ref{sec:system_props} provides analytical results on long-term energy consumption and service rate of the system.  Section~\ref{sec:simulation} includes results from extensive simulated experiments which illustrate the efficacy of our approach, and we conclude in Section~\ref{sec:conclusion}.

\section{System Modeling}\label{sec:modeling}
\vspace{-0.05in}
Consider a single robot in an environment with $n$ communication subsystems, each consisting of a pair of source-destination nodes, as shown in Fig.~\ref{fig:overview}. Data arrives at each source node in a stochastic manner and needs to be communicated to its corresponding destination with a minimum delay. However, the source-destination pairs are located such that direct communication between them is not possible, possibly due to a large distance or blockage from objects in the environment. Thus, a mobile robot is tasked with acting as a relay, constantly planning its path in the area in order to transfer the information from each source to its corresponding destination while minimizing the overall delay. In this section, we present the models for communication, channel prediction, \added{data transfer}, and motion used in this paper, as well as a \added{Markov chain model useful for analyzing long-term properties of the system.}

\vspace{-0.2in}
\subsection{Communication and Channel Prediction Model}\label{sec:modeling.tx}
\vspace{-0.05in}
Communication from a source to its destination involves data transfer across two channels: the link from the source to the \wdh{robot} and the link from the \wdh{robot} to destination. When the source (robot) transmits with power $\Gamma_T$, the received Signal-to-Noise Ratio (SNR) at the \wdh{robot} (destination) is given by $SNR_{\text{rec}} = \Gamma_T\times \Upsilon$, where $\Upsilon$ is the Channel-to-Noise Ratio (CNR). The effects of path loss, shadowing, and multipath fading result in a spatially-varying CNR.

A minimum Bit Error Rate (BER) or other Quality of Service (QoS) requirement induces a minimum received SNR requirement for reliable communication, $SNR_{\text{th}}$. If $SNR\geq SNR_{\text{th}}$, the source (robot) can communicate with the \wdh{robot} (destination). Since the robot and the sources are assumed to have a fixed transmission power, the SNR threshold translates to a minimum required CNR threshold $\Upsilon_{\text{th}}$, and the spatially-varying channels dictate the connectivity regions.

In order for the robot to constantly plan its path and relay the information between the source-destination pairs, it needs to assess the connected region for each communication link, \textit{i.e.}, the connected regions for communication from each source node to itself as well as the connected regions for transmission from itself to each destination. This, however, requires the robot to predict the channel quality at unvisited locations over the workspace. In this paper, we utilize our past work on stochastic channel prediction~\cite{TWC11} to enable the robot to predict its connectivity regions. This approach uses a spatial stochastic process model that accounts for the effects of path loss, shadowing, and multipath fading. More specifically, the channel is characterized by path loss parameters $\theta$, shadowing power $\alpha^2$, shadowing decorrelation distance $\beta$, and multipath fading power $\sigma^2$. Given a very small number of prior channel measurements, the CNR (in dB) at an unvisited location $x$, $\Upsilon_\text{dB}(x)$, can then be best modeled by a Gaussian random variable with its mean and variance given as follows:
\vspace{-0.05in}
\begin{equation}\label{eq: channel prediction}
\fontsize{9.5}{9.5}
\begin{split}
&\mathbb{E}[{\Upsilon}_\text{dB}(x)] = H_x\hat{\theta}+\Psi^T(x)\Phi^{-1}(Y_m-H_m\hat{\theta}),\\
&\Sigma(x) = \hat{\alpha}^2 + \hat{\sigma}^2-\Psi^T(x)\Phi^{-1}\Psi(x),\\[-0.03in]
\end{split}
\vspace{-0.15in}
\end{equation}
where $Y_m = [y_1,\,...,\,y_m]^T$ are the $m$ priorly-collected CNR measurements (in dB), $X_\text{msr} = [x_1^\text{msr},\,...,\,x_m^\text{msr}]$ are the measurement locations, $\hat{\theta}$, $\hat{\alpha}$, $\hat{\beta}$, and $\hat{\sigma}$ are the estimated channel parameters (using $Y_m$), $H_x = [1\ -10\text{log}_{10}(\|x-x_b\|)]$ with $x_b$ denoting either the location of the source in the source-to-\wdh{robot} channel or destination in the \wdh{robot}-to-destination channel, $H_m = [H_{x_1^\text{msr}}^T,\,...,\,H_{x_m^\text{msr}}^T]^T$, $\Psi(x) =
 [\hat{\alpha}^2\text{exp}(-\|x-x_1^\text{msr}\|/\hat{\beta}),\,...,\,\hat{\alpha}^2\text{exp}(-\|x-x_m^\text{msr}\|/\hat{\beta})]^T$, and $\Phi = \Omega + \hat{\sigma}^2 I_m$ with $[\Omega]_{i,j} = \hat{\alpha}^2\text{exp}(-\|x_i^\text{msr}-x_j^\text{msr}\|/\hat{\beta}),\ \forall i,\,j\in \{1,\,...,\,m\}$ and $I_m$ denoting the $m\times m$ identity matrix.

This Gaussian process model allows for the calculation of the probability that the CNR exceeds the minimum CNR imposed by the QoS requirement. Specifically, at point $x$: $\text{P}(\Upsilon_\text{dB}(x) \geq \Upsilon_\text{th, dB}) = \mathbb{Q}((\Upsilon_\text{th, dB} - \mathbb{E}[{\Upsilon}_\text{dB}(x)])/ \sqrt{\Sigma(x)})$, where $\mathbb{Q}(\cdot)$ represents the complementary cumulative distribution function of the standard normal distribution. The prior channel measurements required to make this prediction can be provided by static sensors in the field, gathered in previous operations or at the beginning of the operation, and/or obtained via crowdsourcing.\footnote{The prior channel measurements that the robot can obtain are based on the downlink channel. When it needs to predict the uplink channel, the prior measurements can be collected by the remote sensor which can then send the needed parameters back to the robot for uplink channel prediction.}

The robot can only service the sources from locations where the CNR for both the source-to-robot and robot-to-destination channels exceed the threshold $\Upsilon_{\text{th,dB}}$. We call this the \textit{true relay region}: $R_i^\text{true} = \{x\in\mathbb{R}^2| \Upsilon_{i,s,\text{dB}}^\text{true}(x)\geq \Upsilon_{\text{th,dB}},\, \Upsilon_{i,d,\text{dB}}^\text{true}(x)\geq \Upsilon_{\text{th,dB}}\}$, where $\Upsilon_{i,s,\text{dB}}^\text{true}(x)$ and $\Upsilon_{i,d,\text{dB}}^\text{true}(x)$ are the true CNR in dB at location $x$ of, respectively, the source-to-robot and robot-to-destination channels for the $i^{\text{th}}$ source-destination pair.

The robot, however, will not know $R_i^\text{true}$ and \wdh{must} predict the relay regions \wdh{with} the \wdh{above} channel prediction framework. Assuming independent channels, the probability of a successful communication between the $i^{\text{th}}$ source-destination pair with the \wdh{robot} at position $x$ is
\vspace{-0.09in}
\begin{equation}\label{eq:p_sd}
 p_\text{i,sd}(x)=\text{P}(\Upsilon_{i,s,\text{dB}}(x)\geq \Upsilon_{\text{th,dB}})\times \text{P}(\Upsilon_{i,d,\text{dB}}(x)\geq \Upsilon_{\text{th,dB}}),
 \vspace{-0.02in}
\end{equation}
where $\Upsilon_{i,s,\text{dB}}(x)$ and $\Upsilon_{i,d,\text{dB}}$ are the predicted CNR in dB at location $x$ of, respectively, the source-to-robot and robot-to-destination channels for the $i^{\text{th}}$ source-destination pair. By applying a threshold to the probability of successful end-to-end communication, $p_\text{th}$, we then have a \textit{predicted relay region}: $R_i=\{x\in\mathbb{R}^2| p_{i,\text{sd}}(x) \geq p_\text{th}\}$. \added{Due to the irregular spatial variations of real-world communication channels, these regions will be highly non-convex.}

All links are assumed to be of bandwidth $B\,$Hz, and both the source and robot transmit with \added{a fixed} spectral efficiency of $\xi\,$bps/Hz\added{, \textit{e.g.}, they transmit with a fixed M-QAM constellation}. \added{As a result, when servicing, the robot offloads data from the source at a rate of $\xi\times B$~bps.}

\vspace{-0.2in}
\subsection{\added{Data Accumulation and Offloading}}\label{sec:modeling.AandO}
\vspace{-0.05in}
Let $\{q_1, ...\, q_n\}$ represent infinite-capacity queues at the sources. We assume that the data arrive at $q_i$ according to a Poisson process with average rate $\lambda_i\,$bps. The \textit{traffic} for a single queue is denoted by $\rho_i = \lambda_i\zeta$. For the entire system, we have $\lambda_s = \sum_{i=1}^n \lambda_i$ and $\rho_s = \sum_{i=1}^n \rho_i$.

The \added{\textit{average wait time}, $\bar{W}$, is the average} duration of time between the moment a bit arrives in the queue and the moment it begins to be sent, and the \textit{service time}, $\zeta=1/(\xi B)$, \added{is the time required to transfer a single bit from source to the robot, assuming a successful transmission, which will be part of the optimization framework. Thus, the average total time between the moment a bit enters a queue and the moment it arrives at the destination is $\bar{W}+2\zeta$. The \textit{service discipline} describes how many bits to service during a single visit to a source-destination pair. We consider the exhaustive service policy, in which the robot continues service until the queue is empty. For any robotic routing policy, this service discipline minimizes average wait time~\cite{Liu1992OnOP}.}

\vspace{-0.15in}
\subsection{Motion Model}
\vspace{-0.05in}
We assume that when in motion, the robot travels at a constant velocity, and we model the motion power as a linear function of speed. Specifically, we use a model derived from experimental studies which holds for a large class of robots (\textit{e.g.}, Pioneer robots)~\cite{mei2006deployment}: $\Gamma_m(v) = \kappa_1 v + \kappa_2$ for $v>0$ and $\Gamma_m(v) = 0$ for $v=0$, where $v$ is the robot's speed, and $\kappa_1$ and $\kappa_2$ are positive constants determined by the robot's load and mechanical system. We consider trajectories where, for each pair, the robot chooses a single \textit{relay position}, $r_i\in R_i$, at which the robot stops to relay, so that when traveling between two relay positions $r_i$ and $r_j$, the motion energy consumed is given by $\mathcal{E}_m = (\kappa_1 + \kappa_2/v)||r_i - r_j||$.

\added{The \textit{robotic routing policy} determines the sequence in which the pairs are serviced. As discussed in Section~\ref{sec:introduction}, we are interested in stochastic robotic routing policies characterized by an irreducible Markov chain, $\mathcal{Q}$, with a transition matrix $P$, where $p_{i,j}$ is the probability of servicing the $j^\text{th}$ source-destination pair next, given the robot is currently servicing the $i^\text{th}$ pair. We denote with $\pi=[\pi_1, ..., \pi_n]$ the stationary distribution of $\mathcal{Q}$.}

\added{The \textit{switching time} is the duration of time between the moment the robot completes service for one pair and begins service at the next. Let $S$ denote the symmetric matrix of switching times, with $s_{i,j}$ denoting the switching time between $q_i$ and $q_j$. The switching times are determined by the relay positions and the robot's speed: $s_{i,j} = ||r_i - r_j||/v$. We write $S(X_r)$ to convey the dependency of $S$ on $X_r:=\{r_1, ..., r_n\}$.}
\vspace{-0.1in}
\subsection{\added{Markov Chain Model for Persistent Operation}}\label{sec:modeling.markov}
The instant the robot begins to service a queue is referred to as a \textit{polling instant}, and the duration of the robot's operation may be decomposed into a sequence of \textit{stages}, with the $k^\text{th}$ stage beginning and ending at the $k^\text{th}$ and $(k+1)^\text{th}$ polling instants, respectively. Let $L_k=[L_{1,k}, ..., L_{n,k}]^T$ and $Q_k\in\{q_1, ..., q_n\}$ denote the queue length at each source and the source being polled, respectively, at the $k^{\text{th}}$ polling instant. The sequence of random variables $\mathcal{L}:=\{(L_k,\, Q_k)\}_{k\geq 0}$ forms a Markov chain, and for the robotic routing policy and service discipline discussed above, the chain is stable, \textit{i.e.}, positive recurrent, under the mild assumption that $\rho_s < 1$ and $\mathcal{Q}$ is irreducible~\cite{Takagi1986AnalysisOP}. Thus, empirical time averages of system properties converge to the expected values indicated by the unique stationary distribution.

\added{Analyzing the system via the properties of the Markov chain’s stationary regime assumes that robotic operation is of sufficient duration for system metrics to approach the values predicted by an asymptotic analysis. In practice, this time is typically short.}

\added{The average wait time in the stationary regime is a function of the average arrival rates, service time, switching times, and transition probabilities. Specifically, we have the following lemma:

\begin{lemma}\label{lma:wt_time}
The average wait time $\bar{W}$ in a polling system under the exhaustive service policy and a Markovian routing policy is given by:
\begin{equation}\label{eq:avg_wait_time}
\begin{split}
    \Bar{W}(\lambda, \zeta, S, P)= \underbrace{\frac{\rho_s \zeta}{2(1-\rho_s)}}_{\text{\added{M/G/1 wait time}}} +\quad \underbrace{\frac{1}{2\bar{s}}\sum_{i=1}^n \pi_i \sum_{l=1}^n p_{ij}s_{i,j}^2 + \frac{1}{\bar{s}\rho_s}\sum_{i=1}^n \pi_i \sum_{j=1}^n p_{ij}s_{i,j}\sum_{k\neq i}\rho_k \bar{T}_{ki}}_{\text{\added{additional wait time due to non-zero switching times}}}
\end{split}
\vspace{-0.2in}
\end{equation}
with $\bar{s} = \sum_{i=1}^N \pi_i \sum_{j=1}^N p_{ij}s_{i,j}$ denoting the average switching time and $\bar{T}_{ki}$ denoting the expected time between any departure from $q_i$ and the most recent previous departure from $q_k$. Note that $\bar{T}_{ki}$ is a function of $P$ and can be found by solving a system of $n^2$ equations.
\end{lemma}
\begin{proof}
The proof proceeds by first decomposing total wait time into the wait time of an equivalent $M/G/1$ queue and additional wait time incurred due to switching. This is then followed by a long derivation. See Boxma and Weststrate for details [37]. \cite{Boxma1989WaitingTI},
\end{proof}
}

\added{In the context of our relay system, $\zeta$ is fixed and given by the spectral efficiency and communication bandwidth. Switching times, $S(X_r)$, are determined by the robot's velocity, $v$, and the choice of relay positions, $X_r$. The relay positions are chosen from the relay regions, $R_i$, which are in turn determined by the spatially-varying channel qualities, the QoS requirements, and the transmission power $\Gamma_\text{t}$. The transition probabilities $P$ are precisely the robotic routing policy, while the arrival rates $\lambda_i$ are exogenous. Thus, given the spectral efficiency, $\xi$, channel bandwidth, $B$, the robot's velocity $v$, and the robotic relay policy$(X_r, P)$, we can calculate average wait time in the stationary regime as $\bar{W}(\lambda, 1/(\xi B), S(X_r), P)$ using Lemma~\ref{lma:wt_time}.}

\begin{table*}
    \centering
    \begin{tabular}{|p{20pt}|p{190pt}|p{20pt}|p{190pt}|}
        \hline & & & \\[-0.08in]
            $n$ & Number of source-destination subsystems & $\bar{W}$ & Average wait time of the system\\
            $q_i$ & Queue at the source of the $i^\text{th}$ subsystem & $R_i$ & Predicted relay region of the $i^\text{th}$ subsystem\\
            $r_i$ & Relay position of the $i^\text{th}$ subsystem & $\bar{s}$ & Average switching time \\
             $S(X_r)$ & Matrix of switching times between the relay positions & $\zeta$ & Service time, \textit{i.e.}, time to transmit a single bit from source to destination\\
            $\lambda_i$ & Data arrival rate in bits per second for the $i^\text{th}$ subsystem & $\lambda_s$ & System-wide data arrival rate\\
            $\rho_i$ & Traffic of the $i^\text{th}$ subsystem & $\rho_s$ & System-wide traffic \\
            $P$ & Probability transition matrix for queue visits & $\pi$ & Queue visit frequencies\\
            $\Gamma_m(v)$ & Motion power for velocity $v$ & $\Gamma_t$ & Transmit power \\
            $\mathcal{Q}$ & Markov chain describing the sequence of queues serviced & $\mathcal{L}$ & Markov chain with $\mathcal{L}_k=(L_k, Q_k)$ giving the queue lengths and queue serviced at the $k^\text{th}$ polling instant, respectively.\\
            \hline
        \end{tabular}
        \vspace{-0.05in}
    \caption{List of Key Variables}
    \label{tab:vars_list}
    \vspace{-0.25in}
\end{table*}
\vspace{-0.15in}
\section{Problem Formulation}\label{sec:problem}
\vspace{-0.05in}
Consider our problem of a robot tasked with relaying data
between $n$ source-destination pairs. As mentioned earlier, an
important performance metric of such a robotic relay system is
the average wait time. \added{Therefore, our objective is to minimize the system-wide average wait time\footnote{\added{We note that minimizing the maximum wait time can be another objective of interest, depending on the scenario \cite{avi-itzhak_levy_raz_2008, Shapira2016OnFI}.}} as given in Eq. (3).} Our problem is formally stated as:

\vspace{-0.1in}
\begin{mini}|s|[2]                   
    {X_r,P}                               
    { \Bar{W}&(\lambda, \zeta, S(X_r), P)}   
    {\label{eq:XPopt}}             
    {}                                
    \addConstraint{P }{\in \mathcal{P},\quad X_r\in \prod_{i=1}^n R_i}    
    \vspace{-0.1in}
\end{mini}
where $\mathcal{P}$ is the set of transition matrices describing irreducible Markov chains. 
\added{
\begin{remark}
In Section V, we prove that long-term energy consumption (i.e., stationary values) does not depend on the robotic relay policy. On the other hand, as we shall see, per stage energy consumption \textit{is} a function of the robotic relay policy. Sections IV-A and V-B show how our proposed approach implicitly minimizes average energy per stage.
\end{remark}
}

This problem, however, is challenging to analyze and solve due to the coupling of robotic path planning and data servicing. In fact, simpler versions of this problem, without the robotic component, are still difficult to analyze and have been the subject of extensive studies in both polling systems and queuing theory. We next present special cases of Problem (\ref{eq:XPopt}) which motivate the rest of this paper and further put this optimization problem in the context of existing work.
\vspace{-0.25in}
\subsection{Special Case: No Robotic Position Optimization}\label{sec:problem_ps_equivalence}
\vspace{-0.05in}
In this section, we briefly consider two special cases in which the robotic operation facet of our problem becomes irrelevant, so that (\ref{eq:XPopt}) reduces to problems which have been studied extensively in queuing theory and polling systems. First, if the robot need not move at all, the solution to (\ref{eq:XPopt}) may be found using results on M/G/1 queues, as stated in the following theorem:
\vspace{-0.2in}
\begin{theorem}[M/G/1 equivalence for the case of no motion]\label{thrm:overlapping_rr}
If $\mathcal{I} := \bigcap_{i=1}^n R_i \neq \emptyset$, then the $\{ (X_r^*, P^*)| r_i^*=r_j^*\in \mathcal{I},\, \forall i,j; P\in \mathcal{P} \}$ is the solution set of (\ref{eq:XPopt}), and $ \bar{W}^* = (\rho_s \zeta)/(2(1-\rho_s)$ is the optimal value.
\vspace{-0.05in}
\end{theorem}
\begin{proof} If $\mathcal{I} := \bigcap_{i=1}^n R_i \neq \emptyset$, then all the predicted connected regions are overlapping. As such, the robot can stay in one location to service all source-destination pairs, \textit{i.e.}, there is effectively no need for an unmanned vehicle. In this special case, all switching times become zero. As shown in ~\cite{Boxma1989WaitingTI}, the polling system's wait time is minimized when all switching times are zero. We are thus left with an equivalent M/G/1 queuing system whose average wait time is given in the theorem, regardless of the transition probabilities.
\end{proof}
\vspace{-0.08in}

Second, if each region $R_i$ consists of only a single point, then optimization over relay positions becomes trivial. From the angle of minimizing average wait time, the relay system becomes mathematically equivalent to the polling system studied in \cite{Boxma1989WaitingTI}. No exact algorithm is known for this problem, and minimal wait times are found only through approximations \cite{Boxma1993EfficientVO, Bertsimas1993OptimizationOP}.
\vspace{-0.1in}
\subsection{Coupled Path Planning and Communication Problem}
\vspace{-0.05in}
\added{While the results in Section~\ref{sec:problem_ps_equivalence} help root our work in existing literature, relay position optimization in those special cases is trivial. We now focus on the more realistic instances of Problem~(\ref{eq:XPopt}) in which relay positions must be optimized along with transition probabilities.}

Among the many difficulties presented by this problem is the complex relationship between $\bar{T}_{ki}$ and $P$. To better focus on the new dimension of path planning and avoid repeatedly solving the system of $n^2$ equations that relate $\bar{T}_{ki}$ and $P$, we follow a common approach in polling systems literature~\cite{Boxma1993EfficientVO} and restrict transition probabilities so that $p_{i,j} = p_{k,j} = \pi_j$, $\forall i,j,k\in\{1, ..., n\}$, which we term a \textit{stochastic} robotic routing policy. This permits a closed-form expression for $\bar{T}_{ki}$, as we characterize next:
\vspace{-0.05in}
\begin{theorem}\label{thrm:Tbar_rand}
For a polling system with stochastic routing, the expected time between any departure from $q_i$ and the most recent previous departure from $q_k$ is given by
\vspace{-0.05in}
\begin{equation}\label{eq:tbar_rand}
    \bar{T}_{ki} = \frac{\rho_i \bar{t}}{\pi_i} + \bar{s}_i + \frac{(\rho_s - \rho_k)\bar{t}}{\pi_k} + \frac{1-\pi_k}{\pi_k}\sum_{h=1}^n \pi_h \sum_{l\neq k} \pi_l s_{h,l}\;,
    \vspace{-0.05in}
\end{equation}
where $\bar{t}$ is the average stage duration (mathematically characterized in Lemma~\ref{lma:tbar} of Section~\ref{sec:system_props}), and $\bar{s}_i$ is the first moments of the switching time from any relay position to $r_i$.
\vspace{-0.05in}
\end{theorem}
\begin{proof}
See Appendix for the proof.
\end{proof}
\vspace{-0.05in}

Stochastic robotic routing policies still allow for arbitrary visit frequency and security benefits while reducing the complexity of the design space. Robotic relay policies thus consist of the tuple $(X_r,\, \pi)$, and (\ref{eq:XPopt}) can be restated as:
\vspace{-0.05in}
\begin{mini}|s|[2]                   
    {X_r,\pi}                               
    {&\Bar{W}(\lambda, \zeta, S(X_r), P\added{(\pi)})\quad\quad\quad\quad\quad\quad}   
    {\label{eq:XPopt_rand}}             
    {}                                
    \addConstraint{\sum_{i=1}^n \pi_i=1,\quad}{\pi_i> 0\;,\forall i,\quad X_r}{\in \prod_{i=1}^n R_i \quad.}
    \vspace{-0.17in}
\end{mini}
\added{where $P(\pi)$ indicates the dependence of $P$ on $\pi$.}

We refer to $\pi_i$ as the \textit{visit frequency} of $q_i$. We are then interested in finding the visit frequencies and relay positions that minimize average wait time, which is simpler than the original problem in~(\ref{eq:XPopt}). This problem, however, is still quite challenging and lacks structure that would make it easily solvable. In the next section, we then show how to find approximately-optimal solutions.

\vspace{-0.1in}
\section{Approximately-Optimal Robotic Relay Policies}\label{sec:AORP}
\vspace{-0.05in}
The optimization problem in (\ref{eq:XPopt_rand}) is intractable due to the highly non-convex nature of the relay regions $R_i$ and the strong coupling between $X_r$ and $\pi$. Thus, it is necessary to find \textit{approximately}-optimal robotic relay policies (AORP's). To reduce complexity, we propose an iterative approach which minimizes the average switching time over $X_r$ with $\pi$ fixed, then minimizes $\bar{W}$ over $\pi$ with $X_r$ fixed. This may be repeated until convergence. \wdh{Algorithm~\ref{alg:AORP} outlines the approach.} We denote the variables associated with the solution to Algorithm~\ref{alg:AORP} with the AORP superscript, so the visit frequencies, robotic relay positions, and average wait time corresponding to the AORP are given by $\pi^\text{AORP}$, $X_r^\text{AORP}$, and $\bar{W}^\text{AORP}$, respectively. We next discuss the two stages of this approach in detail.
\vspace{-0.1in}
\begin{algorithm}
\SetAlgoLined
\KwResult{$\pi^\text{AORP}$, $X_r^\text{AORP}$}
 For each source-destination pair:\\
 \textbf{Step 1}: Find the predicted relay regions ($R_i$), using the channel prediction model of Section~\ref{sec:modeling.tx}.\\
 \textbf{Step 2}: Find $a_i$, the $\alpha$-shape of $R_i$.\\
 \textbf{Step 3}: Simplify $a_i$ using Ramer-Douglas-Peucker algorithm.\\
 \textbf{Step 4}: Find convex partition of $a_i$ using Hertel-Mehlhorn algorithm.\\
 \vspace{0.05in}
 Initialize $\pi^\text{AORP}$ with Eq.~\ref{eq:sqrt_pi}. Then iteratively optimize $\pi^\text{AORP}$ and $X_r^\text{AORP}$:\\
 \textbf{Step 5}: Solve Problem~\ref{eq:MISOCP} to update $X_r^\text{AORP}$.\\
 \textbf{Step 6}: Solve Problem~\ref{eq:PIopt} to update $\pi^\text{AORP}$.\\
 \textbf{Step 7}: Repeat steps 5 and 6 until convergence.
 \caption{Approximately-Optimal Robotic Relay Policy (AORP)}
 \label{alg:AORP}
 
\end{algorithm}
\vspace{-0.2in}
\subsection{Optimization of Robotic Path Planning}\label{sec:AORP.position}
\vspace{-0.075in}
In the first part of our iterative approach, we fix $\pi$ and find the optimal robotic relay positions $X_r$, simplifying (\ref{eq:XPopt_rand}) to:
\vspace{-0.05in}
\begin{mini}|s|[2]
    {X_r}
    { \Bar{W}&(\lambda, \zeta, S(X_r), \added{P(\pi)})}
    {\label{eq:Xopt_rand}}
    {}
    \addConstraint{X_r}{\in \prod_{i=1}^n R_i}\quad,
    \vspace{-0.12in}
\end{mini}
In general, this problem is non-convex and intractable. In this section, we then show how we can tackle this problem. We first note that relay positions, $X_r$, affect the average wait time, $\Bar{W}$, through the switching times, $s_{i,j}$. Specifically, compared to the wait time of the system with no switching times (\textit{i.e.}, an M/G/1 queue), the average additional wait time due to switching times is positive everywhere except when the average switching time is zero ($\bar{s}=0$), as discussed in the proof to Theorem~\ref{thrm:overlapping_rr}. Therefore, an intuitive heuristic simplification for solving (\ref{eq:Xopt_rand}) is to minimize the average switching time instead. Furthermore, minimizing the average switching time amounts to minimizing the average distance traveled per stage of operation, which is a common objective in robotics. \added{In fact, as we show in Section~\ref{sec:system_props.energy}, this is equivalent to minimizing average energy consumed per stage.} Thus, rather than solving (\ref{eq:Xopt_rand}), we find $X_r$ that minimizes the average switching time, as follows: 
\vspace{-0.04in}
\begin{mini!}|s|[2]
    {X_r,S}
    { \sum_{i=1}^n \pi_i \sum_{j=i}^n \pi_j \;s_{i,j} \label{eq:min_avg_switch_obj}}
    {\label{eq:min_avg_switch}}
    {}
    \addConstraint{||r_i - r_j||_2 =}{vs_{i,j},\; \forall\; i,\,j\label{eq:min_avg_switch_xs_to_ds}}
    \addConstraint{X_r}{\in \prod_{i=1}^n R_i\label{eq:min_avg_switch_r_in_X}}\quad.
    \vspace{-0.45in}
\end{mini!}
The objective is linear, and if (\ref{eq:min_avg_switch_xs_to_ds}) is relaxed, \textit{i.e.}, $||r_i - r_j||_2 \leq vs_{i,j}$, it becomes a second-order cone constraint which will be active at the optimal solution. We next show how to partition the regions $R_i$ so that (\ref{eq:min_avg_switch_r_in_X}) can be stated as a combination of linear and integer constraints, making the problem a mixed-integer second-order cone program (MISOCP).

\subsubsection{Relay Region Partitioning}\label{sec:AORP.position.partition}
\begin{figure*}
    \centering
    \includegraphics[width=6in, height=1.6in]{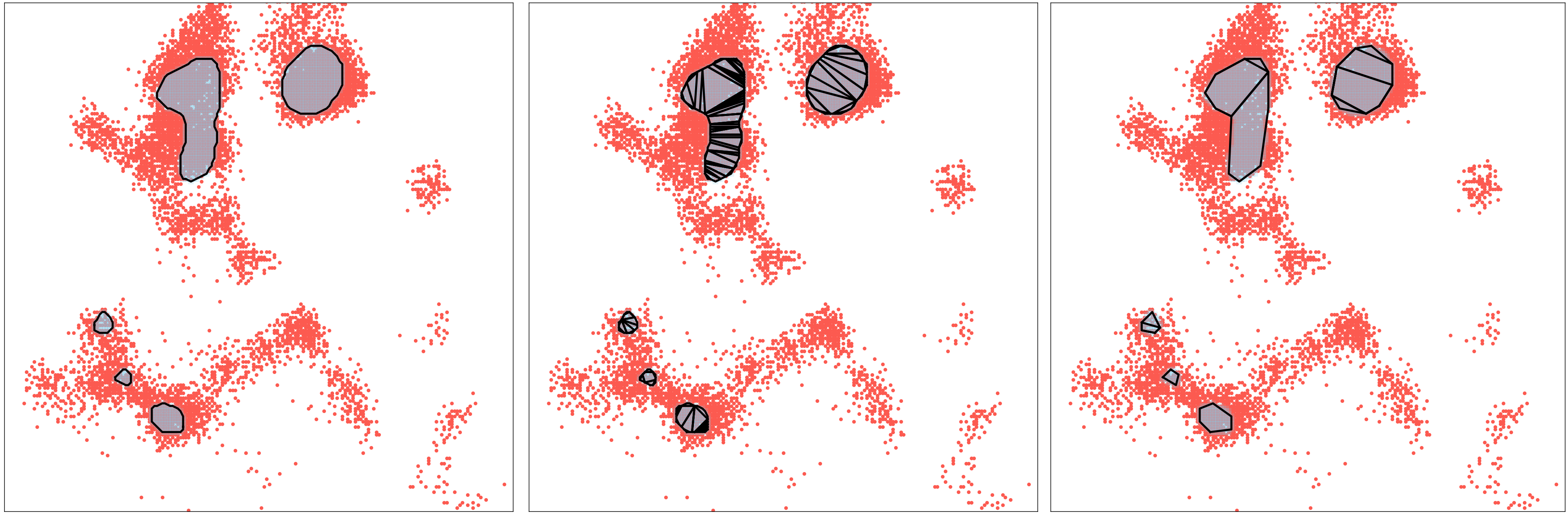}
    \caption{Extraction of a convex partition of the predicted relay region. First, evaluating the probability of connectivity across a fine grid produces a set of points $R_1$ at which $p_{1,\text{sd}} > p_\text{th}$. Here $p_\text{th}=0.7$, and $R_1$ is shown shaded in grey on top of the true relay region, shown in red. (Left) Extracting the $\alpha$-shape of $R_1$ produces several non-convex polygons, 5 in this case. Then the resulting polygons are partitioned into convex regions using the Hertel-Mehlhorn algorithm. (Center) Without RDP smoothing, this produces 113 convex polygons. (Right) Using RDP to smooth before performing the partition reduces the number of polygons to 11. See the color pdf for better viewing.}
    \label{fig:decomp_stylized}
    \vspace{-0.25in}
\end{figure*}

In real channel environments, each relay region and predicted relay region can be irregular and non-convex in its shape. However, there are many clear advantages to optimizing over well defined regions in $\mathbb{R}^2$. Thus, to solve (\ref{eq:min_avg_switch}) efficiently, we next propose a procedure, illustrated in Fig.~\ref{fig:decomp_stylized}, which converts each set $R_i$ into a set of convex polygons $C_i=\{C_1^i, ..., C_{m_i}^i\}$ whose union approximates the predicted relay region $R_i$.\footnote{As seen in Section~\ref{sec:simulation}, each $R_i$ is not necessarily a joint set.}

\begin{enumerate}[label=\roman*,leftmargin=1.5\parindent]
    \item Convert the points $R_i$ to a set of (possibly non-convex and non-simple) polygons. This may be achieved by finding the $\alpha$-shape of the points, as originally proposed in~\cite{Edelsbrunner1983AS}. The $\alpha$-shape of a set of points is a set of polygons that best fit the original set, where the value of $\alpha$ is chosen based on the granularity of the grid to make the resulting polygons as tight to the points as desired.

    \item Smooth the non-convex polygons using the Ramer-Douglas-Peucker (RDP) algorithm (\cite{Ramer1972AnIP,Douglas1973ALGORITHMSFT}). This will reduce the number of convex polygons required for partitioning. Tolerance on the amount of area lost or added by smoothing is set by the corresponding RDP parameters.

    \item Finally, perform convex partitioning on each of the smoothed polygons. A number of algorithms exist for such a partitioning. We elect to use the Hertel-Mehlhorn algorithm~\cite{Hertel1983FastTO}, which has a time complexity of $O(n)$ and is guaranteed to produce at most four times the minimal number of convex polygons required for the partition.
    \vspace{-0.02in}
\end{enumerate}
This approach produces a reasonable number of convex partitions that fit tightly to the predicted relay region, as illustrated in Fig.~\ref{fig:decomp_stylized}. \wdh{The} procedure \wdh{is} performed only once for each source-destination pair, and the same partitions are then used in each iteration of Algorithm~\ref{alg:AORP}.

\subsubsection{Mixed Integer Second-Order Cone Problem Formulation}
With the convex partitions, we can now reformulate (\ref{eq:min_avg_switch}) as a MISOCP. First, note that for each convex polygon $C_k^i$, there exists $A_k^i$, $b_k^i$, such that $A_k^i r_i - b_k^i\leq \mathbf{0}_n$ if and only if $r_i$ is in $C_k^i$, where $\mathbf{0}_n$ indicates a vector of length $n$ of all 0's, and with an abuse of notation, `$\leq$' is the entry-wise comparison. Furthermore, we introduce a set of indicator variables $\eta_k^i$, where $\eta_k^i = 1$ if $r_i\in C_k^i$ and $\eta_k^i = 0$ otherwise, and $\eta$ is the collection of all $\eta_k^i$. Then, we have the following mixed-integer program (MIP):
\begin{mini!}|s|[2]                   
    {X_r,S,\eta}                               
    {\sum_{i=1}^n \pi_i \sum_{j=i}^n \pi_j \;s_{i,j} \label{eq:MIQCP_OB}}   
    {\label{eq:MIQCP}}             
    {}                                
    \addConstraint{||r_i - r_j||_2 \leq}{vs_{i,j},\; \forall\; i,\,j\label{eq:MIQCP_soc}}
    \addConstraint{\eta_k^i (A_k^i r_i - b_k^i) }{\leq \mathbf{0}_n,\;\forall i, k \label{eq:MIQCP_regions}}
    \addConstraint{\sum_{k=1}^{m_i} \eta_k^i}{ = 1,\;\forall i,\quad \eta_k^i \in\{0,1\}, \;\forall i, k \label{eq:MIQCP_etas}\quad.}
    \vspace{-0.12in}
\end{mini!}
where constraints (\ref{eq:MIQCP_regions}) and (\ref{eq:MIQCP_etas}) guarantee $r_i\in R_i$. However, the quadratic constraint (\ref{eq:MIQCP_regions}) contains both continuous and integer variables, making the problem intractable for many MIP solvers. We next show how to make (\ref{eq:MIQCP_regions}) linear.

Let $r_{i,1}$ and $r_{i,2}$ denote the first and second coordinates of $r_i$, respectively, and let $R_{i,k,\text{min}}=\text{min} \{r_{i,k}\,|\, r_i \in R_i\}$ and $R_{i,k,\text{max}} = \text{max} \{r_{i,k}\,|\, r_i \in R_i\}$. First, we constrain $r_i$ to be within the bounding box $\mathbb{B}_i=[R_{i,1,\text{min}}, R_{i,1,\text{max}}]\times [R_{i,2,\text{min}}, R_{i,2,\text{max}}]$. Then for some large constant $c_k^i>\max_{r_i\in B_i}(||A_k^i r_i||_\infty)$, let $\tilde{b}_k^i = b_k^i+c_k^i\mathbf{1}_n$, $\tilde{A}_k^i = [A_k^i,\, c_k^i\mathbf{1}_n]$, and $\tilde{r}_k^i = [r_i,\, \eta_k^i]^T$, with $\mathbf{1}_n$ a vector of 1's with length $n$. Then $\tilde{A}_k^i \tilde{x}_k^i - \tilde{b}_k^i \leq \mathbf{0}_n$ if and only if $\eta_k^i (A_k^i r_i - b_k^i) \leq \mathbf{0}_n$, and the problem can be rewritten as a MISOCP:
\vspace{-0.08in}
\begin{mini!}|s|[2]                   
    {X_r,S,\eta}                               
    { \sum_{i=1}^n \pi_i \sum_{j=i}^n \pi_j \;s_{i,j} \label{eq:MISOCP_OB}}   
    {\label{eq:MISOCP}}             
    {}                                
    \addConstraint{||r_i - r_j||_2 \leq}{vs_{i,j},\; \forall\; i,\,j\label{eq:MISOCP_soc}}
    \addConstraint{\tilde{A}_k^i \tilde{r}_k^i - \tilde{b}_k^i \leq \mathbf{0}_n,\;\forall i, k}{,\quad r_i\in \mathbb{B}_i,\;\forall i\label{eq:MISOCP_regions_lin}}
    \addConstraint{\sum_{k=1}^{m_i} \eta_k^i}{ = 1,\;\forall i,\quad \eta_k^i \in\{0,1\}, \;\forall i, k \label{eq:MISOCP_etas}\quad.}
    \vspace{-0.1in}
\end{mini!}
The constraints in (\ref{eq:MISOCP_regions_lin}) are now linear, so this problem can be solved with mathematical solvers which provide guarantees of optimality and finite time convergence.
\vspace{-0.1in}
\subsection{Optimizing the Stochastic Robotic Routing Policy}
\vspace{-0.05in}
In the second part of our iterative procedure, we fix $X_r$ to focus on finding the optimum stochastic robotic routing policy $\pi$. As a result, the optimization problem of (\ref{eq:XPopt_rand}) simplifies to:
\vspace{-0.075in}
\begin{mini}|s|[2]
    {\pi}
    { \Bar{W}(\lambda, \zeta, S(X_r), \added{P(}\pi\added{)})}
    {\label{eq:PIopt}}
    {}
    \addConstraint{\sum_{i=1}^n \pi_i=1,}{\quad \pi_i> 0,\;\forall i\quad.}
    \vspace{-0.1in}
\end{mini}
This problem is mathematically similar to the stochastic polling system optimization problem, and as discussed in Section~\ref{sec:problem_ps_equivalence}, no exact algorithm is known for solving it in its general form. Thus, similar to~\cite{Boxma1993EfficientVO}, we use successive quadratic approximations to find approximate solutions. Specifically, we use sequential least squares programming (SLSQP), which finds a local minimum. We next briefly discuss a special case from polling systems literature for which a closed form solution to (\ref{eq:PIopt}) does exist, and discuss its potential relevance to our robotic routing problem.
\subsubsection{Square Root Rule Approximation}
For a polling system with exhaustive service, Poisson arrivals, and a stochastic robotic routing policy, if all switching times are identical, then average wait time becomes convex in $\pi$, and a closed-form solution exists for the optimal visit frequencies, as stated in the following lemma:
\vspace{-0.05in}
\begin{lemma}\label{lma:sqrrt_rule}
Let $\pi^*$ be the visit frequencies which minimize the average wait time in a polling system  with exhaustive service, Poisson arrivals, and stochastic routing. If $s_{i,j} = s_{k,l},\,\,\forall i,j,k,l\in\{1, ..., n\}$, then
\vspace{-0.05in}
\begin{equation}\label{eq:sqrt_pi}
    \pi_i^* = \frac{\sqrt{\rho_i(1-\rho_i)}}{\sum_{j=1}^n\sqrt{\rho_j(1-\rho_j)}}\quad.
\end{equation}
\vspace{-0.1in}
\end{lemma}
\begin{proof}
See~\cite{Boxma1993EfficientVO}.
\vspace{-0.1in}
\end{proof}

While Eq.~(\ref{eq:sqrt_pi}) results in a closed-form solution, it requires identical switching times. In our robotic routing context, this will only hold if all switching times are zero since $s_{i,i}=0$. This then results in the trivial case discussed in Theorem~\ref{thrm:overlapping_rr}. As such, this special case does not directly apply to our scenario. However, we have observed from extensive simulations that Eq.~(\ref{eq:sqrt_pi}) is a good approximation when all switching times except the self-loops, $s_{i,i}$, are similar, \textit{i.e.}, when $s_{i,j}\approx s_{k,l}$, $\forall i,j,k,l\in\{1, ..., n\},\; i\neq j, k\neq l$. In other words, Eq.~(\ref{eq:sqrt_pi}) may be a good approximation if the optimum robotic relay positions become approximately equal distanced. We leave rigorous investigation of this approximation to future work.\footnote{Note that even if we assume there is some form of delay that results in non-zero $s_{i,i}$'s, such values would be negligible as compared to the time it takes for the robot to travel from one relay region to the next, which is what differentiates this problem from polling systems.}

\subsubsection{Observed Transition Probabilities}\label{sec:AORP.fv.obs}
The visit frequencies $\pi$ found by solving (\ref{eq:PIopt}) are with respect to the discrete time Markov chain produced by taking snapshots of the system at polling instances, i.e., $\mathcal{Q}$. Consider the robot immediately after it completes servicing source $q_i$. With probability $\pi_i$, the robot then transitions back to servicing $q_i$. However, under the exhaustive service policy adopted in this paper (see Section~\ref{sec:modeling.AandO}), $q_i$ will be empty at this point, resulting in the robot eventually finding another source to service instantly (i.e., no time expires). As such, it makes sense to exclude the self loops for simplicity, as summarized next:
\vspace{-0.075in}
\begin{remark}
The observed stochastic robotic routing policy is Markovian with transition probabilities $\tilde{P}$ given by
\begin{equation}
    \tilde{p}_{ij} = \begin{cases}
        \frac{\pi_j}{\sum_{k\neq i}\pi_k} & \text{if } j\neq i,\\
        0 & \text{otherwise.}
    \end{cases}
\end{equation}
\end{remark}
\vspace{-0.04in}
Consequently, the robot is always observed leaving a relay position immediately upon service completion, and the observed visit frequencies, $\tilde{\pi}$, may be found by finding the stationary distribution associated with $\tilde{P}$, i.e., $\tilde{\pi}^T\tilde{P} = \tilde{\pi}^T$. These observed transition probabilities give clearer descriptions of the robotic relay's behavior and are used in Section~\ref{sec:simulation} to analyze the results of the simulated experiments.
\vspace{-0.15in}
\subsection{A Derived Deterministic Robotic Relay Policy}\label{sec:AORP.AORPT}
\vspace{-0.05in}
After convergence, an additional step may be taken to produce a routing table policy based on the AORP in which the relay positions and visit frequencies are the same but the robot visits the pairs according to a deterministic, periodic (not cyclic) sequence determined by $\pi^\text{AORP}$, using the Golden Ratio rule~\cite{Itai1984AGR}. We refer to this policy as the AORP-Table (AORPT). The AORPT policy eliminates the need to select the next source to service on the fly while ensuring the same visit frequencies. As the traditional state-of-the-art in robotics route design is also deterministic, we use the AORPT for comparison in Section~\ref{sec:simulation}. We next give a brief example.

Consider the case with four source-destination pairs with $\pi=[1/2, 1/4, 1/8, 1/8]$. The Golden Ratio Rule \wdh{produces} the following visit sequence for the robot \wdh{to follow repeatedly}:
\vspace{-0.1in}
\begin{equation}\label{eq:gold_seq}
(q_1, q_2, q_1, q_3, q_1, q_2, q_1, q_4)\quad.
\vspace{-0.1in}
\end{equation}
The resulting policy is periodic but not cyclic. Furthermore, it spaces visits evenly throughout the period. To illustrate, consider an alternative sequence which satisfies the visit frequencies $\pi$:
\vspace{-0.1in}
\begin{equation}
(q_1, q_2, q_1, q_2, q_1, q_3, q_1, q_4)\quad.
\vspace{-0.1in}
\end{equation}
Here the two visits to $q_2$ both occur in the first half of the sequence whereas in (\ref{eq:gold_seq}), the visits are evenly distributed.

In summary, in this section, we showed how to find approximately-optimal solutions to the optimization problem in (\ref{eq:XPopt_rand}). Our proposed approach first finds an efficient convex partition for each predicted relay region. It then iteratively optimizes the robotic relay positions to minimize average switching time while fixing the robotic routing policy, followed by optimizing average wait time over the routing policy (\textit{i.e.}, visit frequencies) while fixing the relay positions. After convergence, the visit frequencies can further be translated to a deterministic policy for efficient robotic operation in the field.
\vspace{-0.15in}
\section{\added{Long-Term Average Power Consumption and Service Rate}}\label{sec:system_props}
\vspace{-0.05in}
In this section, we present important system properties regarding the \added{long-term average power consumption and service rate}. The results further motivate the choice used for the optimization problem formulation of the past sections. Our analysis focuses on average values when operating in the stationary regime. Thus, the results hold both in expectation across an ensemble of realizations over a finite number of stages, with initial conditions drawn according to the stationary distribution of the Markov chain $\mathcal{L}$, and in averaging asymptotically over time for any realization of the process, regardless of initial conditions. \added{We first prove that the percentage of time spent servicing is independent of the robotic relay policy $(X_r, P)$. We then use this finding to derive key results regarding average power consumption and service rate.}
\vspace{-0.1in}
\subsection{Relating Percentage of Time Servicing to System Traffic}
\vspace{-0.05in}
We first examine the significance of the system traffic, $\rho_s$, more closely and show that it gives the long-term percentage of time the robot spends servicing bits in a queue. Recall that, as presented in Section~\ref{sec:modeling.markov}, robotic operation may be decomposed into a series of stages demarcated by the polling instants, and that $\mathcal{L}$ is a Markov chain with $\mathcal{L}_k$ giving the state of the system, \textit{i.e.}, the queue lengths and current queue under service, at the $k^\text{th}$ polling instant. Recall further that the robotic relay system is stable in the sense that $\mathcal{L}$ is positive recurrent if and only if $\rho_s<1$ and $\mathcal{Q}$ is irreducible. We are now ready to present the following lemma:
\vspace{-0.05in}
\begin{lemma}\label{lma:tbar}
For any stable robotic relay policy, average stage duration in the stationary regime, $\bar{t}$, is given by $\bar{t}=\bar{s}/(1-\rho_s)$, where $\bar{s}$ is the average switching time.
\vspace{-0.05in}
\end{lemma}
\begin{proof}
Under the assumption of stability, the average number of bits entering the system in a stage, $\bar{t}\lambda_s$, equals the average number leaving, $(\bar{t} - \bar{s})/\beta$. The result follows directly.
\vspace{-0.05in}
\end{proof}

As $\bar{s}$ is determined by our robotic relay policy, the average stage duration and the average time spent servicing during a single stage are likewise determined by the policy. However, Lemma~\ref{lma:tbar} also indicates that the proportion $\bar{s}/\bar{t}$ is fixed (equal to $\rho_s$) regardless of robotic relay policy, and similarly, the long-term percentage of time spent servicing is independent of the policy, as stated in the following theorem. Here and throughout the rest of the section, we use `$\;\,\aseq\;$' to indicate almost sure equality, and $t_o$ is the time of operation.
\vspace{-0.05in}
\begin{theorem}\label{thrm:rho_s}
Under any stable robotic routing policy, the system traffic $\rho_s$ gives the long-term percentage of time spent servicing, \textit{i.e.}, $ \lim_{t_o\to\infty} \phi(t_o)/t_o\; \aseq\; \rho_s$,
where $\phi(t_o)$ is the total time spent servicing during operation.
\vspace{-0.05in}
\end{theorem}
\begin{proof}
$\phi(t_o)$ may be decomposed into the sum of all service times completed during the full stages and an additional service time completed for the current, incomplete stage. $t_o$ may be decomposed similarly so that
\vspace{-0.05in}
\begin{equation}
\begin{split}
    \lim_{t_o\to\infty} \frac{\phi(t_o)}{t_o} &= \lim_{t_o\to\infty}\frac{\left(\sum_{k=1}^{K(t_o)} \phi_k + \phi_\epsilon\right)/K(t_o)}{\left(\sum_{k=1}^{K(t_o)} t_k + t_\epsilon\right)/K(t_o)}\aseq\; \lim_{K\to\infty} \frac{\left(\sum_{k=1}^{K} \phi_k\right)/K}{\left(\sum_{k=1}^{K} t_k \right)/K}\;\aseq\;\frac{(\bar{t} - \bar{s})}{\bar{t}} = \rho_s\;,
\end{split}
\vspace{-0.07in}
\end{equation}
where $K(t_o)$ is the number of complete stages after operation of duration $t_o$, $\phi_k$ is the time spent servicing during the $k^\text{th}$ stage, $\phi_\epsilon$ is any additional service time, $t_k$ is the duration of the $k^\text{th}$ stage, and $t_\epsilon$ is any additional time. The second equality holds because $t_\epsilon$ and $\phi_\epsilon$ are almost surely finite and the remaining equalities hold due to the positive recurrence of $\mathcal{L}$ as guaranteed by the assumption of stability.
\end{proof}
\vspace{-0.05in}
\vspace{-0.2in}

\subsection{Energy Consumption \added{and Average Power}}\label{sec:system_props.energy}
\vspace{-0.05in}
Consider average energy consumption over a single stage of operation, including both communication and motion energy. We have the following theorem:
\vspace{-0.05in}
\begin{theorem}\label{thrm:e_per_st}
For any stable robotic relay policy, the average energy consumed over a single stage while operating in the stationary regime, $\bar{\mathcal{E}}_\text{st}$, is given by $\bar{\mathcal{E}}_\text{st} =\bar{t}(\Gamma_m(v)(1-\rho_s) +  \Gamma_t\rho_s)$
\vspace{-0.05in}
\end{theorem}
\begin{proof}
Note that $\bar{\mathcal{E}}_\text{st} = \bar{s}\,\Gamma_m(v) + (\bar{t} - \bar{s})\Gamma_t$. The result follows directly \wdh{from Lemma~\ref{lma:tbar}}.
\vspace{-0.05in}
\end{proof}
Recalling from Lemma~\ref{lma:tbar} that $\bar{t}$ is a linear function of $\bar{s}$, we see the energy \textit{per stage} is a linear function of average switching time $\bar{s}$, which is in turn determined by the robotic relay policy (see Section~\ref{sec:modeling.AandO}). In other words, the optimization problem in (\ref{eq:Xopt_rand}) amounts to minimizing the average energy consumed per stage, which is an important metric in robotics literature. We next \added{move from stage-level analysis to} formally consider the long-term average power consumption of the robot.		
\vspace{-0.07in}
\begin{theorem}\label{thrm:AP}
Let $\mathcal{E}(t_o)$ be the energy consumed during operation of duration $t_o$. Then under any stable robotic relay policy $\lim_{t_o\to\infty} \mathcal{E}(t_o)/t_o\;=\;\; \aseq \;\Gamma_m(v)(1-\rho_s) +  \Gamma_t\rho_s$.
\vspace{-0.05in}
\end{theorem}
\begin{proof}
Note that $\mathcal{E}(t_o) = (t_o-\phi(t_o))\Gamma_m(v) + \phi(t_o)\Gamma_t$, where the first and second terms give the motion and communication energy, respectively. The result then follows from Theorem~\ref{thrm:rho_s}.
\vspace{-0.05in}
\end{proof}
				
While Theorem~\ref{thrm:AP} is an asymptotic result, in practice it holds for a long enough period of time (corroborated by simulation results of Section~\ref{sec:simulation}), as summarized next:																					
\vspace{-0.05in}																									
\begin{remark}\label{rmk:long_run_energy}
For operation over a sufficiently large but finite duration $t_o$, $\mathcal{E}(t_o)$ is well approximated by $\mathcal{E}(t_o) \approx (\Gamma_m(v)(1-\rho_s) +  \Gamma_t\rho_s) t_o.$
\end{remark}
\vspace{-0.2in}

\subsection{System Long-Term Average Service Rate}
\vspace{-0.05in}
The average number of bits serviced during a stage must equal the average number of bits that enter the system for any stable policy (see Lemma~\ref{lma:tbar}), which leads to the following: 
\vspace{-0.075in}
\begin{theorem}\label{thrm:MB}
Under any stable robotic relay policy, $\lim_{t_o\to\infty} d(t_o)/t_o\; \aseq\; \lambda_s$, where $d(t_o)$ is the total number of bits serviced during operation.
\vspace{-0.05in}
\end{theorem}
\begin{proof}
From the assumption of stability, data in and data out must be equal in the long-term.
\end{proof}
\vspace{-0.05in}

While Theorem~\ref{thrm:MB} is an asymptotic result, we find in practice that the empirical service rate converges quickly to $\lambda_s$, which leads to the following remark:
\vspace{-0.05in}
\begin{remark}
The total bits serviced during operation over a sufficiently large but finite duration $t_o$ is well approximated by $d_\text{out}(t_o) \approx \lambda_s t_o$.
\vspace{-0.05in}
\end{remark}

Importantly, systems with identical average long-term service rates may have vastly different average wait times. As a trivial example, consider two systems which each service a single bit during operation over a fixed amount of time. The first services the bit immediately and then idles the remainder of the time, while the second idles for a time and then services the bit at the very end of operation. While the average service rate of the two systems is identical, the first system results in less wait time. As such, the average wait time is the key metric to consider when optimizing robotic operation.

\section{Simulation Results}\label{sec:simulation}
\vspace{-0.05in}
In this section, we demonstrate the performance and efficacy of our approach with extensive simulations in realistic channel environments. First, we discuss channel prediction and the extraction of convex partitions from the predicted relay regions. We then extensively study the performance of our approach by showing the effect of various system parameters on the AORP. We further show that our proposed approach can significantly reduce the average wait time compared to the state-of-the-art. Finally, we discuss the average service rate and energy consumption of the AORP to corroborate Theorems~\ref{thrm:MB} and~\ref{thrm:AP}.

In our implementation of Algorithm~\ref{alg:AORP}, (\ref{eq:XPopt_rand}) is solved using the SLSQP solver of SciPy 1.7.1, and (\ref{eq:MISOCP}) is solved with IBM CPLEX 12.9. To initialize $\pi$, we use the approximation in (\ref{eq:sqrt_pi}), and in all simulations, we keep the following system parameters fixed: channel bandwidth $B=2\,$Mhz, spectral efficiency $\xi = 8\,$bits/s/Hz, robot's constant velocity $v = 1\,$m/s, and motion power parameters $\kappa_1 = 7.2\,$N and $\kappa_2 = 0.29\,$W.
\vspace{-0.2in}

\subsection{Relay Region Prediction}\label{sec:simulation.predict}
\vspace{-0.05in}
We first generate the source-robot and robot-destination communication channels using realistic channel parameters derived from real-world measurements in downtown San Francisco. The following channel parameters, introduced in Section~\ref{sec:modeling.tx}, are extracted from the real channel measurements of~\cite{smith2004urban}: $\alpha^2=16$, $\beta=2.09\,$m, $\theta = [5.2\,\text{dB}, -7.5]^T$ and $\sigma^2=1.96$. Then, different channel instances are generated based on these parameters using the probabilistic channel generator described in \cite{JRobotics11}. Briefly, \cite{JRobotics11} utilizes a given set of path loss, shadowing and multipath parameters to generate a 2D channel whose underlying parameters match the given set. The robot and sources transmit with a power of $\Gamma_t = 100\,$mW, the receiver noise power is $-85\,$dBm, and the acceptable SNR threshold for all the channels is set to $33\,$dB. Given the transmit power, the SNR threshold translates to a CNR threshold of $\Upsilon_\text{th, dB} = 13\,$dBm. This then dictates the true relay regions, which the robot does not know but needs to predict.

The robot then uses $1\%$ apriori channel samples for each channel to predict the channel over the \wdh{rest} of the workspace, which is unvisited. \wdh{U}sing the approach outlined in Section~\ref{sec:modeling.tx}\wdh{,} the robot \wdh{predicts} the relay regions by calculating the probability of successful end-to-end communication, $p_{i,\text{sd}}$, based on the CNR threshold (see Eqs.~(\ref{eq: channel prediction}) and (\ref{eq:p_sd})).
\vspace{-0.2in}

\subsection{Relay Region Convex Partitioning}\label{sec:simulation.decomp}
\begin{figure}
    \centering
    \raisebox{-0.5\height}{\includegraphics[trim= 0 0 0 0, clip, width=6.5in]{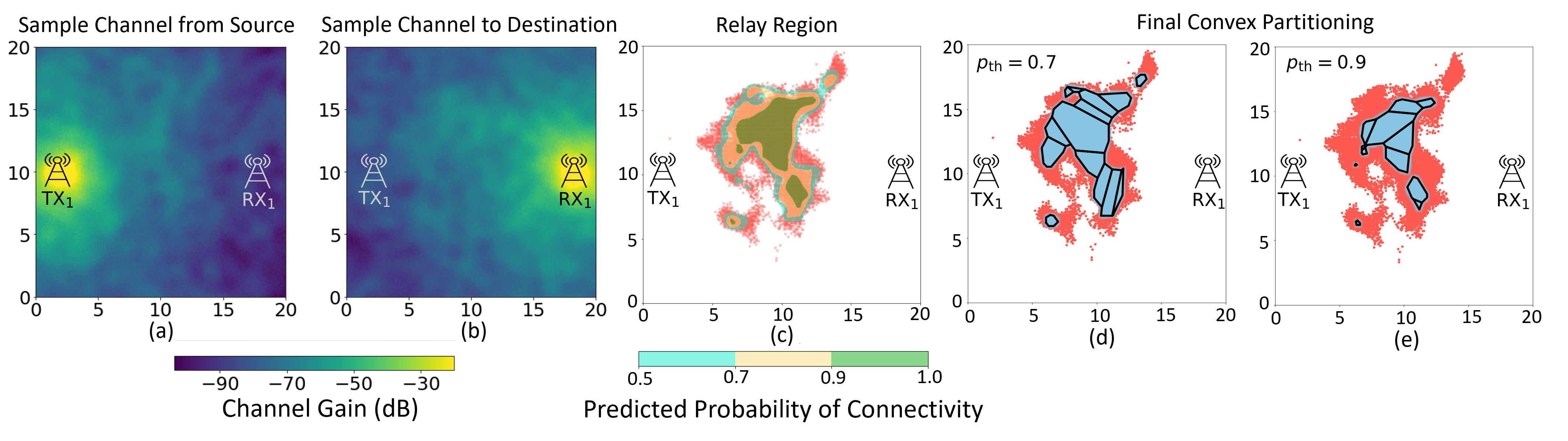}}

    \raisebox{-0.5\height}{\footnotesize
     \begin{tabular}{|c|c|c|c|}
        \hline
             & $p_\text{th}=0.5$&$p_\text{th}=0.7$ & $p_\text{th}=0.9$ \\
        \hline
             $p_c$ &  0.84 & 0.93 & 0.99\\
        \hline
            $|C|$ & 15 & 14 & 12\\
        \hline
    \end{tabular}}
   \caption{\added{Sample generated (a) source-to-robot and (b) robot-to-destination channels. The spatial variation of the channel is highly non-convex, as can be seen. (c) True relay region (red) and contours of predicted probability of successful end-to-end communication $p_{\text{sd}}$. (d,e) Final convex partitioning given two threshold probabilities $p_{\text{th}}$. The predicted relay regions (blue) are overlaid on the true relay region (red). (Bottom) The table shows $p_c$, the probability that a point in the predicted connected region is truly connected, and $|C|$, the number of convex partitions, for three values of $p_\text{th}$.} See the color pdf for better viewing.}
 \label{fig:decomp_comparison}
 \vspace{-0.28in}
\end{figure}

\vspace{-0.05in}
We next show the process of extracting the convex partitions as proposed in Section~\ref{sec:AORP.position.partition}, which plays a central role in the optimization of robotic relay positions. \added{Fig. \ref{fig:decomp_comparison} (a,b) shows an example of source-to-robot and robot-to-destination channels, generated with the aforementioned real-world parameters.} The robot then uses $1\,$\% a priori channel samples and predicts the channel at the rest of the workspace in order to calculate $p_{i,\text{sd}}$ across a fine grid of points. We construct \wdh{the relay region} by selecting those points for which $p_{i,\text{sd}}\geq p_\text{th}$. Fig.~\ref{fig:decomp_comparison} (c-e) shows a true connected region and two samples of the predicted regions for two values of $p_\text{th}$. As can be seen, for $p_\text{th}=0.9$, the prediction framework is more conservative in declaring a location as connected, resulting in a smaller predicted relay region. As expected, fewer convex partitions are required for $p_\text{th}=0.9$. 

The table then compares the probability of true connectivity as well as the number of associated convex partitions for three values of $p_\text{th}$. As can be seen, the predicted relay regions provide good approximations of the regions where the robot may successfully \added{communicate with both the source and destination}, \textit{i.e.}, $R_i^\text{true}$. In particular, we see that the probability that a point in $R_i$ is in $R_i^\text{true}$ becomes higher as the threshold increases, as expected. We furthermore observe that the probability of true connectivity is larger than the threshold probability, \textit{i.e.}, for $p_c=\text{P}(x\in R_i^\text{true} | x\in R_i)$, we have $p_c > p_\text{th}$. Finally, the number of convex partitions reduces as $p_\text{th}$ increases, as expected.

Overall, as $p\text{th}$ increases, the chance that locations in the predicted relay region are truly connected goes up, and the number of convex polygons required to partition the region decreases, simplifying optimization problem (\ref{eq:MISOCP}). However, increasing $p_\text{th}$ makes the channel prediction more conservative and can lead to inefficient robotic operation as larger areas of the true relay region are excluded from the predicted region (see Fig.~\ref{fig:decomp_comparison} (d,e)). As a result, the robot travels longer distances than needed between relay positions. In the rest of the paper, we use $p_\text{th}=0.7$ for the prediction of all relay regions.

\vspace{-0.2in}

\subsection{Impact of System Parameters}
\vspace{-0.05in}
This section explores the impact of several system parameters on our proposed approach. We start with a system of three source-destination pairs in order to show the impact of several different system parameters on the performance. We then increase the complexity and show the performance for a system consisting of six source-destination pairs.

\subsubsection{Traffic}
\begin{figure*}
    \centering
    \includegraphics[width=6.5in]{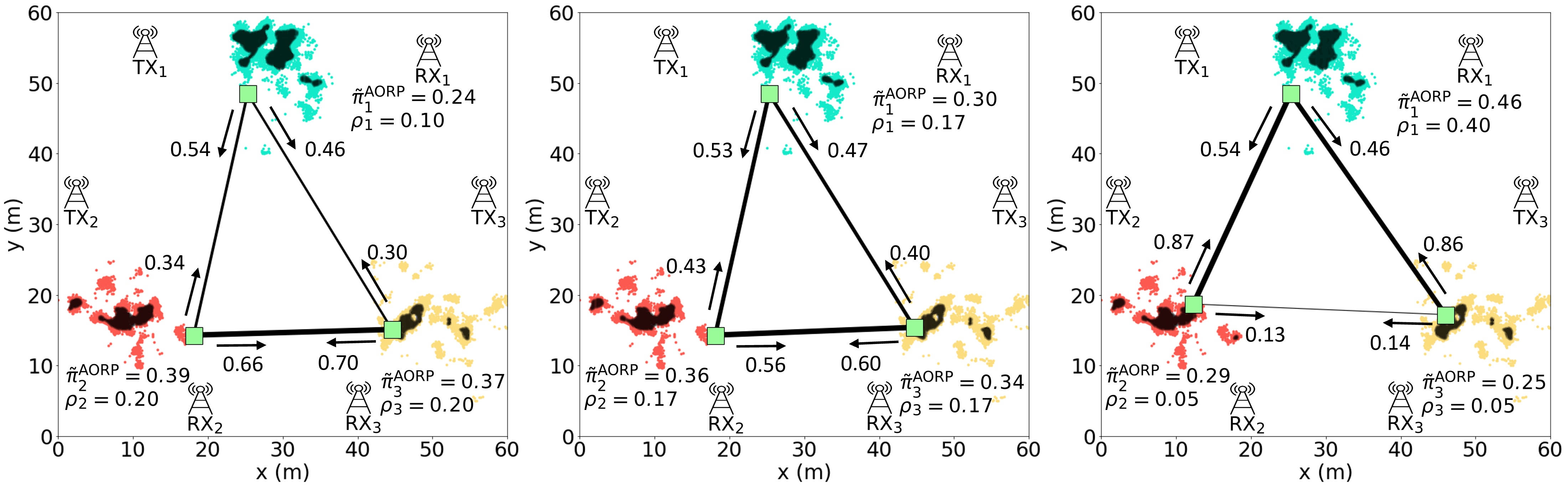}
    \vspace{-0.4in}
    \caption{Effect of traffic distribution on the AORP probability transitions and relay positions. The green squares indicate the relay positions, and the observed transition probabilities, $\tilde{p}_{i,j}^\text{AORP}$, are labeled on the arrows along each edge while the thickness of each edge corresponds to $\tilde{p}_{i,j}^\text{AORP}$ as well. Each subsystem is labeled with both the corresponding incoming traffic, $\rho_i$, and the resulting visit frequency, $\tilde{\pi}_i^\text{AORP}$. For each subsystem, the predicted relay region (in black) is overlaid on the true connected region. Moving from left to right, the increase in $\rho_1$ is reflected in the AORP transition probabilities and visit frequencies, as can be seen. See the color pdf for optimum viewing of this figure.}
    \label{fig:var_traffic}
    \vspace{-0.2in}
\end{figure*}

To isolate the effect of traffic distribution, consider the three-queue system with symmetric placement of the subsystems shown in Fig.~\ref{fig:var_traffic} (we note that the relay regions' irregularity always introduces some asymmetry). In all three cases, system traffic is held constant at $\rho_s = 0.5$ while $\rho_1$ is varied, where $\rho_i$ is the traffic at the $i^\text{th}$ source (see Section~\ref{sec:modeling.AandO}). The remaining traffic is split evenly between $\rho_2$ and $\rho_3$. The AORP transition probabilities are shown on the routes, observed visit frequencies and traffic are labeled on each subsystem, and the relay position within each predicted relay region is marked.

Intuitively, as the traffic $\rho_i$ at $q_i$ increases, $\tilde{\pi}_i^\text{AORP}$ should increase. As can be seen, with $\rho_1 = 0.1$, $q_1$ receives only half the traffic received by either $q_2$ or $q_3$. Then, under the AORP, the robot mainly moves back and forth between $q_2$ and $q_3$, visiting $q_1$ on average only once every four visits. When increasing $\rho_1$ so that the traffic of each queue is $0.17$, $\tilde{\pi}_1^\text{AORP}$ increases so that the robot services each queue with approximately equal frequency. Finally, when $\rho_1=0.4$, $q_1$ receives four times the combined traffic of $q_2$ and $q_3$. As can be seen, about half the visits are made to $q_1$ while the remaining are split between $q_2$ and $q_3$. Furthermore, the relay positions $r_2^\text{AORP}$ and $r_3^\text{AORP}$ move farther away from each other to be closer to $r_1^\text{AORP}$ since the robot will rarely switch between $q_2$ and $q_3$. In summary, when a queue accounts for a greater proportion of the total system traffic, the visit frequency to that queue increases, and the relay positions may shift to account for certain switches occurring more frequently. 

\begin{figure*}
    \centering
    \includegraphics[width=6.5in]{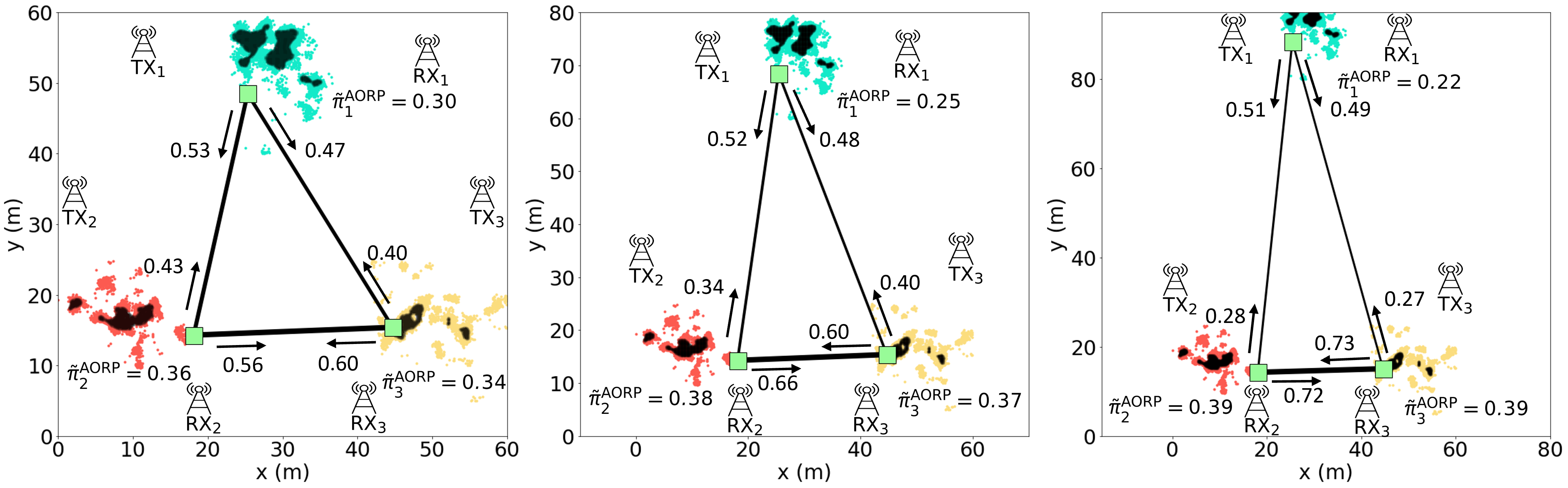}
    \vspace{-0.4in}
    \caption{Effect of subsystem placement on the AORP probability transitions and relay positions. The green squares indicate the relay positions, and the observed transition probabilities, $\tilde{p}_{i,j}^\text{AORP}$, are labeled on the arrows along each edge while the thickness of each edge corresponds to $\tilde{p}_{i,j}^\text{AORP}$ as well. In all cases, the traffic at each queue is $\rho_i=0.17$. For each subsystem, the predicted relay region (in black) is overlaid on the true connected region. Moving from left to right, subsystem one is moved farther away from the rest of the subsystems, and the impact of its increased isolation is then reflected in the AORP transition probabilities and visit frequencies, as can be seen. See the color pdf for optimum viewing of this figure.}
    \label{fig:var_placement}
    \vspace{-0.2in}
\end{figure*}

\subsubsection{Subsystem Placement}
The relative placement of each source-destination subsystem also greatly impacts \wdh{the AORP}, as shown in Fig.~\ref{fig:var_placement}. To isolate the effect of subsystem placement, we keep the traffic at all subsystems equal. The subsystems are first placed so that midpoints between each source-destination pair form the vertices of an equilateral triangle. The first subsystem is then increasingly offset in the direction of the positive $y$-axis, as shown.

When the placement is symmetric (Fig.~\ref{fig:var_placement} (left)), each queue is serviced with approximately the same frequency. The slight difference in visit frequencies can be explained by the particular shape of the predicted relay regions, which are naturally different from each other. In other words, although the source-destination pairs are positioned symmetrically, different shapes and locations of the predicted relay regions result in $r_2^\text{AORP}$ and $r_3^\text{AORP}$ being placed closer to one another than either can be to $r_1^\text{AORP}$, thus making the switching time to $q_1$ relatively longer. As a result, $q_1$ is visited less frequently. 

As the first subsystem is offset by first $20\,$m and then $40\,$m in the subsequent figures, $\tilde{\pi}_1^\text{AORP}$ drops to $0.25$, then $0.22$, respectively. Intuitively, the optimum switching times $s_{2,1}$, $s_{1,2}$, $s_{3,1}$ and $s_{1,3}$ increase as $q_1$ moves farther away, so visits to $q_1$ lead to greater wait times in $q_2$ and $q_3$. Thus, $q_1$ is visited less often as it becomes more isolated.

\subsubsection{Impact of Both Traffic and Subsystem Placement}
\begin{figure}
    \centering
    \includegraphics[width=3.5in]{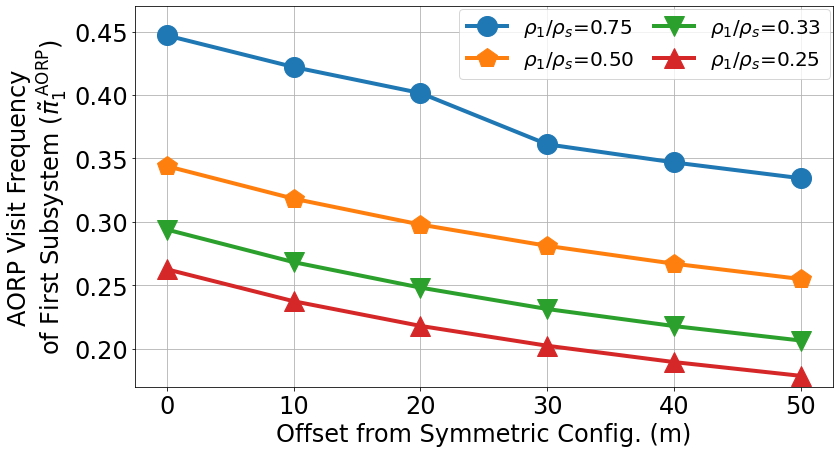}
    \vspace{-0.05in}
    \caption{Impact of subsystem placement on the observed visit frequency of the first subsystem under the AORP ($\tilde{\pi}_1^\text{AORP}$) for different traffic values, \textit{i.e.}, for various values of the ratio $\rho_1/\rho_s$ in the three-queue system. See the color pdf for optimum viewing of this figure.}
    \label{fig:sim2_trends}
    \vspace{-0.2in}
\end{figure}

Fig.~\ref{fig:sim2_trends} shows the impact of both subsystem placement and varying traffic. An offset of $0$ indicates the system is configured symmetrically, as in Fig.~\ref{fig:var_placement} (left), and increasing the offset moves the $q_1$ subsystem away from the original configuration, as shown in Fig.~\ref{fig:var_placement} (center, right). The overall system traffic $\rho_s$ is constant across all experiments, and $\rho_1/\rho_s$ gives the percentage of incoming traffic through $q_1$, which is varying. The remaining traffic is split evenly between $q_2$ and $q_3$ so that traffic is symmetric when $\rho_1/\rho_s=0.33\,$. The figure then shows the AORP observed visit frequency of the first subsystem ($\tilde{\pi}_1^\text{AORP}$). As can be seen, regardless of the distribution of traffic, moving subsystem one farther away, decreases $\tilde{\pi}_1^\text{AORP}$. Likewise, when less traffic arrives at $q_1$, $\tilde{\pi}_1^\text{AORP}$ decreases regardless of offset.										

\subsubsection{Velocity}
\begin{figure}
    \vspace{-0.1in}
    \centering
    \includegraphics[width=6.5in]{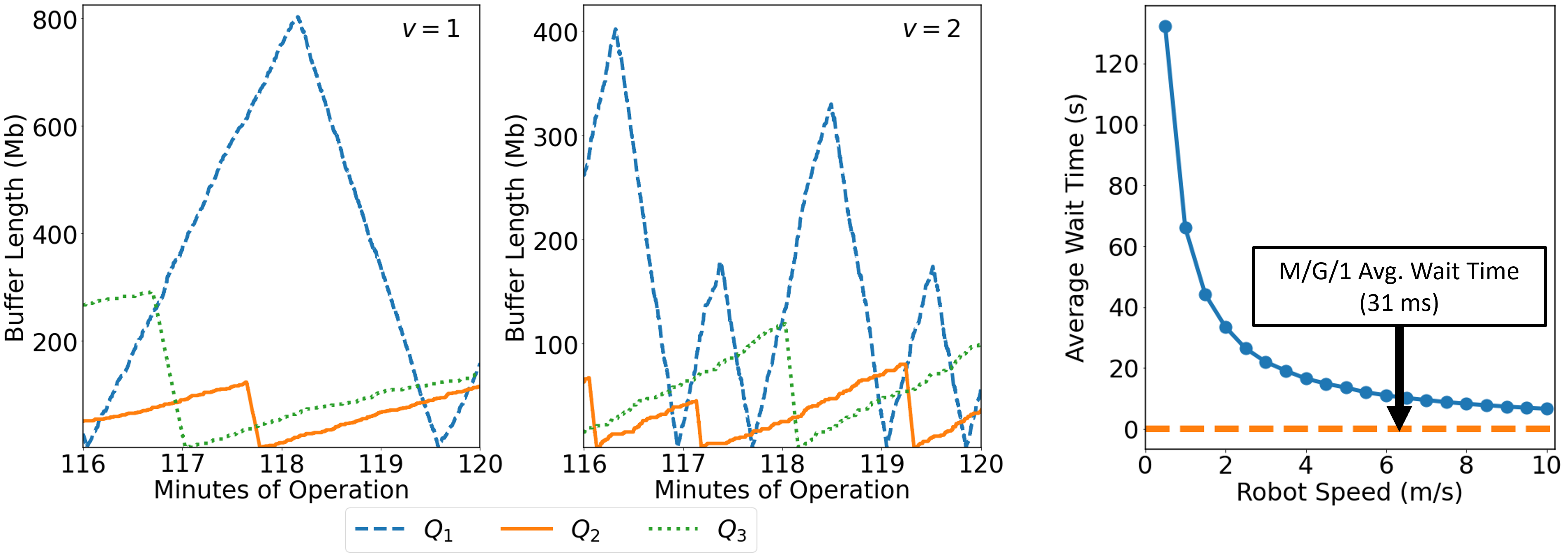}
    \caption{Impact of robot velocity on wait times and queue lengths in a three-queue system. (Left, center) Sample realization of queue lengths during the last four minutes of a two-hour operation period for two sample velocities. The peaks (troughs) indicate polling (switching) instants. Larger $v$ results in shorter stages and queue lengths. (Right) Increasing $v$ also reduces overall average waiting time at a rate of $1/v$. See the color pdf for optimum viewing of this figure.}
            \label{fig:eq_tri_vel_trends}
    \vspace{-0.2in}
\end{figure}

We next discuss the impact of the robot's speed on the performance of the AORP. Switching times scale linearly with the inverse of robot's velocity, \textit{i.e.}, $1/v$, and as such, $v$ significantly impacts the average wait time of the AORP. Specifically, Eqs.~(\ref{eq:avg_wait_time}) and~(\ref{eq:tbar_rand}) indicate that wait time $\bar{W}$ is an affine function of $1/v$ for any stochastic policy. This is seen in Fig.~\ref{fig:eq_tri_vel_trends} (right), which shows the average wait time of the AORP as a function of the robot's speed in the three-queue system shown in Fig.~\ref{fig:var_traffic} (right). Interestingly, as $v\to \infty$, $s_{i,j}\to 0$, $\forall i,j\in \{1, ..., n\}$, and consequently $\bar{W}$ converges to the wait time of an M/G/1 queue (see Theorem~\ref{thrm:overlapping_rr}).

Fig.~\ref{fig:eq_tri_vel_trends} further shows the queue lengths over the last four minutes of a two-hour operation for $v = 1\,$m/s (left) and $v = 2\,$m/s (right). Doubling the velocity reduces all switching times by half, and as this shortens average queue lengths, the time spent servicing during a single stage decreases, too, as mathematically characterized in Section~\ref{sec:system_props}. This results in the completion of approximately twice as many stages (\textit{i.e.}, polling instances), as can be seen.

\subsubsection{A More Complex System}
\begin{figure}
    \centering
    \raisebox{-0.5\height}{\includegraphics[width=4in]{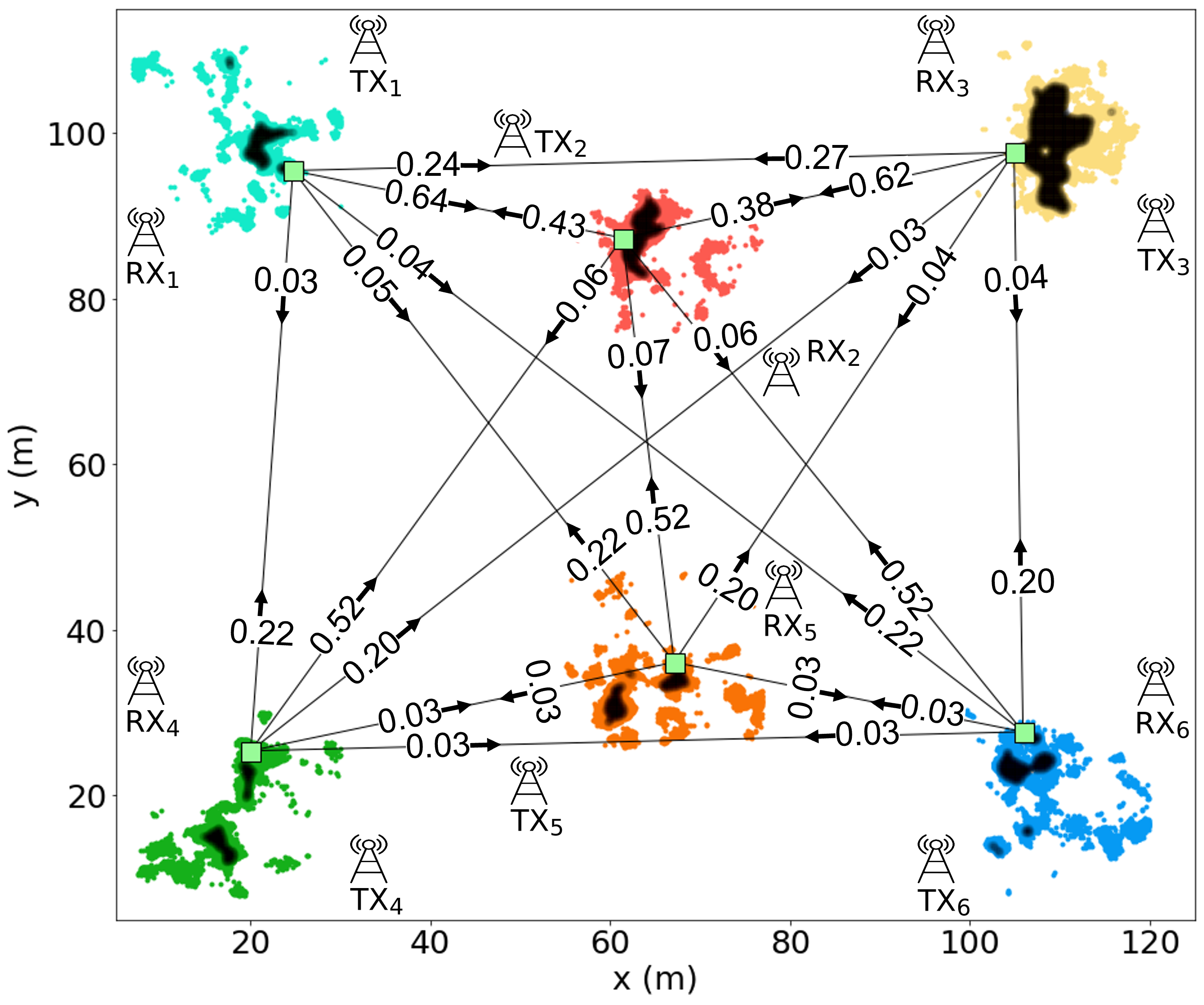}}
    \raisebox{-0.2\height}{\footnotesize
    \begin{tabular}{|c|c|c|}
    \hline
         Subsys. & $\rho_i/\rho_s$ & $\tilde{\pi}_i^\text{AORP}$ \\
         \hline
         1& 0.32& 0.21\\
         \hline
         2& 0.32& 0.50\\
         \hline
         3& 0.32& 0.19\\
         \hline
         4& 0.01& 0.03\\
         \hline
         5& 0.01& 0.04\\
         \hline
         6& 0.01& 0.03\\
         \hline
    \end{tabular}}
    
    \caption{(Left) The AORP for a six-queue system. For each subsystem, the predicted relay region (in black) is overlaid on the true connected region. The transmitter and receiver locations for each source-destination pair are marked, and green squares indicate the relay positions. The observed transition probabilities $\tilde{p}_{i,j}^\text{AORP}$ are labeled along each route. (Right) The table gives the percentage of traffic, $\rho_i/\rho_s$, and observed visit frequency, $\tilde{\pi}_i^\text{AORP}$, for each queue. See the color pdf for optimum viewing.}
    \label{fig:outlier_tspnp_v_aorp}
    \vspace{-0.25in}
\end{figure}

To illustrate how the proposed AORP works in more complex scenarios, we next consider the system of six source-destination pairs shown in Fig. \ref{fig:outlier_tspnp_v_aorp}. The system consists of three high traffic queues ($q_1$, $q_2$, and $q_3$) and three low traffic queues ($q_4$, $q_5$, and $q_6$). Furthermore, the second and fifth subsystems are placed towards the center. 

\added{The table shows} that while the relative traffic at each queue is somewhat reflected in the AORP visit frequencies, they are not exactly proportional due to different locations of the subsystems as well as the underlying space-varying channel quality. The three high traffic queues are visited most, and the edges traversed most frequently form a triangle with vertices given by the relay positions $r_1^\text{AORP}$, $r_2^\text{AORP}$, and $r_3^\text{AORP}$. Among the high (low) traffic queues, $q_2$ ($q_5)$ is visited most often as it is located more centrally compared to $q_1$ and $q_3$ ($q_4$ and $q_6$). Consequently, when the robot moves from $q_1$ to $q_3$, it will frequently stop at $q_2$ along the way. Furthermore, the particular channel realizations in the first, second, and third subsystems results in the second relay position, $r_2^\text{AORP}$, being placed closer to $r_1^\text{AORP}$ than $r_3^\text{AORP}$. As a result, $q_1$ is visited more frequently than $q_3$ despite their symmetry in traffic and subsystem placement. Thus, we again see that the AORP accounts for the differing arrival rates, the relative positioning of the subsystems, and the specific shape of the communication channels, even in more complex systems.

\vspace{-0.15in}
\subsection{Comparison to the State-of-the-Art}

We next compare our proposed robotic routing policy with a baseline approach that would be consistent with the state-of-the-art. To the best of our knowledge, the problem of interest to this paper is solved neither in the robotics nor in the communication literature, so we will use the solution proposed for a similar problem in the literature as a baseline. \added{Specifically, visiting different relay regions in an efficient manner resembles a category of robotic path planning problems known as traveling salesperson problems with neighborhoods, which continues to be the basis of proposed solutions to several data gathering problems in robotics literature \cite{VFDTN,Tsilomitrou2018MobileRT, Peters2019UAVSU, celik2010, Hari2019TheGP, Hari2021OptimalUR, Huang2015SystemTD}}. We then compare our approach with this baseline.

\begin{figure}
    \centering
    \includegraphics[trim=0 0 0 0, clip, width=4in]{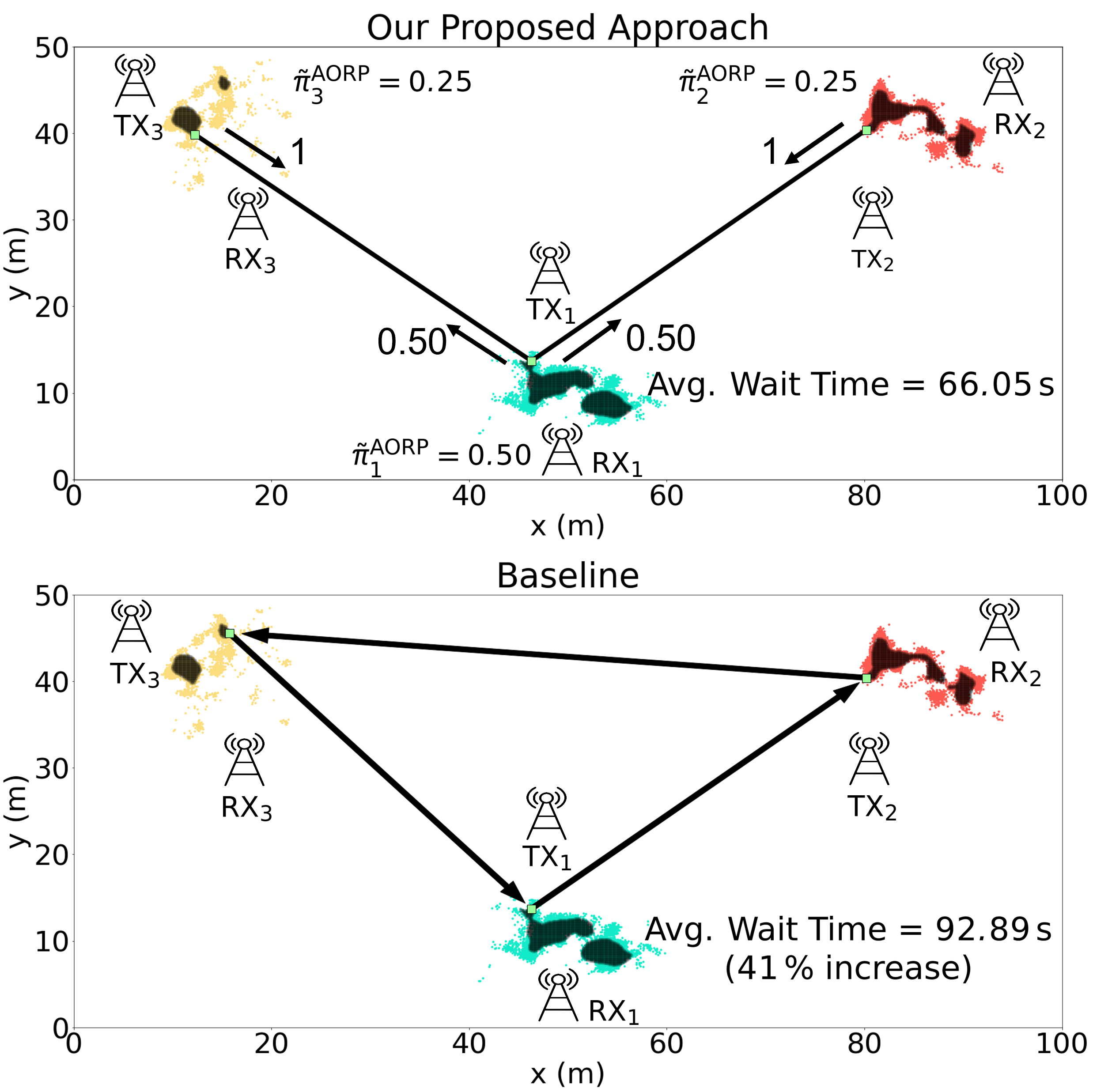}

\vspace{-0.1in}
    \caption{Comparison of our proposed approach (top) and the baseline (bottom) for a three-pair system with $\rho = 0.32$ and $\rho_2 = \rho_3= 0.04$. The observed probability transitions under the AORP, $\tilde{p}_{i,j}$, are labeled along each edge of the AORP, and the thickness of the line between relay positions indicates the relative frequency with which the edge is traversed. Under the AORP, $\tilde{p}_{2,3}\approx\tilde{p}_{3,2} \ll 0.01$, so for ease of presentation, we treat those probabilities as 0. See the color pdf for optimum viewing.}
    \label{fig:asym_tri_AROP_v_TSPNP}
    \vspace{-0.3in}
\end{figure}

Consider the three-queue system shown in Fig. \ref{fig:asym_tri_AROP_v_TSPNP}, with $\rho_1=0.32$, $\rho_2=0.04$, and $\rho_3 = 0.04$. Since the baseline is a deterministic policy, we compare it with AORP-Table as discussed in Section~\ref{sec:AORP.AORPT}. For each policy, we simulate twenty two-hour operation periods with the queues initially empty. As shown in Fig.~\ref{fig:asym_tri_AROP_v_TSPNP}, the wait time under the baseline policy is $92.89\,$s, while under the AORP, it is $66.05\,$s. Thus the average wait time under the baseline policy is $26.84\,$s ($41\,$\%) longer than the average wait time under the AORP. This is achieved through a proper design that jointly considers the path planning and communication aspects of the problem, as we have proposed in this paper. We note that given the conservative bias of our prediction framework as discussed earlier, the probability of the relay positions being truly connected is very high. As such, the relay positions are assumed connected for both our approach and the baseline in the analysis of Fig.~\ref{fig:asym_tri_AROP_v_TSPNP}.

\vspace{-0.15in}
\subsection{Confirmation of the Theories of Section~\ref{sec:system_props}}
\vspace{-0.05in}
Theorems~\ref{thrm:AP} and \ref{thrm:MB} characterized the robot's average service rate and average operation power under a stable policy. We next confirm these theorems with our simulated data. Consider the three-queue system of Fig.~\ref{fig:asym_tri_AROP_v_TSPNP}. Again, we run twenty two-hour simulations, and for each simulation, record the average power and service throughput. \wdh{The average} power \wdh{across all simulations} is $4.58\,$W, consistent with the $4.53\,$W predicted by Theorem~\ref{thrm:AP}. Similarly, the average service rate from simulated data is $6.34\,$Mb/s, while Theorem~\ref{thrm:MB} predicts it to be $6.40\,$mb/s.

As discussed in Section~\ref{sec:system_props}, the overall average power and average service rate are independent of the robotic routing policy (assuming the policy is stable) when averaged over several stages of operation. To this end, we further note that the average power and service rate under the baseline policy were $4.61\,$W and $6.29\,$Mb/s, respectively, which is consistent with the theorems and remarks of Section~\ref{sec:system_props}. As such, the average wait time is the right metric to consider when designing robotic relay policies, as we have done in this paper.
\vspace{-0.1in}

\section{Conclusions}\label{sec:conclusion}
\vspace{-0.05in}
This paper considered a mobile robot (ground or UAV) \wdh{which relays data} between pairs of otherwise far-away source and destination communication nodes. We posed the problem of finding the optimal stochastic robotic relay policy, consisting of visiting frequencies and relay positions. To find approximate solutions to this problem, we proposed a novel algorithm (AORP) which alternately optimizes average wait time over the visiting frequencies and average switching time over relay positions. To minimize the average switching time, we showed how to decompose the highly non-convex relay regions into efficient convex partitions, allowing us to formulate the problem as a MISOCP. Additionally, we mathematically characterized a number of important properties of the system related to the robot’s long-term energy consumption and service
rate. Through extensive simulations with real channel parameters, we showed how various system parameters affect the AORP and further compared with the state-of-the-art.
\vspace{-0.1in}


%
\section*{A\textsc{ppendix} - Proof of Theorem~\ref{thrm:Tbar_rand}}
\vspace{-0.05in}
\begin{proof}
We consider each term in the expression
\vspace{-0.05in}
\begin{equation*}
   \frac{\rho_i \bar{t}}{\pi_i} + \bar{s}_i + \frac{(\rho_s - \rho_k)\bar{t}}{\pi_k} + \frac{1-\pi_k}{\pi_k}\sum_{h=1}^n \pi_h \sum_{l\neq k} \pi_l s_{h,l}\;.
   \vspace{-0.05in}
\end{equation*}
Looking back in time, the process first serviced the data in $q_i$. From \cite{Kleinrock1988TheAO}, the expected time to do so is $\rho_i \bar{t}/\pi_i$ (first term). Immediately preceding this, the robot switched to $q_i$ (second term). With probability $\pi_k$, the queue serviced immediately prior to $q_i$ was $q_k$. With probability $1-\pi_k$, the queue just prior to $q_i$ was not $q_k$, and we expect $1/\pi_k$ additional backward steps before reaching $q_k$. These steps consist of servicing and switching intervals. None of the servicing intervals occur at $q_k$, so the expected number of times $q_j$, $j\neq k$, is serviced is $[\pi_j/(1-\pi_k)]/\pi_k$. Summing over all $j\neq k$, the contribution to $\bar{T}_{ki}$ from these intervals is the third term. Similarly, we expect to have switched away from $q_j$ a total of $\pi_j/\pi_k$ times, but none of those switches will be to $q_k$. Thus the contribution to $\bar{T}_{ki}$ from all these switching times is the fourth term.
\end{proof}




\ifCLASSOPTIONcaptionsoff
  \newpage
\fi


\bibliographystyle{IEEEtran}
{\footnotesize
\bibliography{main}}

\end{document}